\documentclass[10pt,twocolumn,letterpaper]{article}

\usepackage{iccv}
\usepackage{times}
\usepackage{epsfig}
\usepackage{graphicx}
\usepackage{amsmath}
\usepackage{amssymb}

\usepackage{booktabs}
\usepackage{multirow}
\usepackage{algorithm}
\usepackage{algorithmic}
\usepackage{listings}
\usepackage[table,xcdraw,usenames,dvipsnames]{xcolor}
\usepackage{etoolbox}
\usepackage{csquotes}
\usepackage[normalem]{ulem}
\makeatletter
\AfterEndEnvironment{algorithm}{\let\@algcomment\relax}
\AtEndEnvironment{algorithm}{\kern2pt\hrule\relax\vskip3pt\@algcomment}
\let\@algcomment\relax
\newcommand\algcomment[1]{\def\@algcomment{\footnotesize#1}}
\renewcommand\fs@ruled{\def\@fs@cfont{\bfseries}\let\@fs@capt\floatc@ruled
  \def\@fs@pre{\hrule height.8pt depth0pt \kern2pt}%
  \def\@fs@post{}%
  \def\@fs@mid{\kern2pt\hrule\kern2pt}%
  \let\@fs@iftopcapt\iftrue}
\makeatother

\usepackage[pagebackref=true,breaklinks=true,letterpaper=true,colorlinks,bookmarks=false]{hyperref}
\usepackage{enumerate}

\def\Jing#1{{\color{magenta}{\bf [Jing:} {\it{#1}}{\bf ]}}}
\def\YC#1{{\color{blue}{\bf [Yuchao:} {\it{#1}}{\bf ]}}}
\def\Mochu#1{{\color{orange}{\bf [Mochu:} {\it{#1}}{\bf ]}}}
\def\NB#1{{\color{green}{\bf [NB:} {\it{#1}}{\bf ]}}}

\def\rf{\cellcolor[HTML]{B4C7E7}} 
\def\rs{\cellcolor[HTML]{D2DDF1}} 
\def\rt{\cellcolor[HTML]{E9EEF8}} 
\def\tr#1{\multirow{2}{*}{#1}} 
\def\b#1{\textbf{#1}} 
\def\wl#1{\multicolumn{1}{c}{\begin{tabular}[c]{@{}c@{}}{#1}\end{tabular}}} 

\iccvfinalcopy 

\def\iccvPaperID{5676} 
\def\httilde{\mbox{\tt\raisebox{-.5ex}{\symbol{126}}}}

\ificcvfinal\fi

\begin{document}

\title{Measuring and Modeling Uncertainty Degree for Monocular Depth Estimation}

\author{Mochu Xiang\\
Northwestern Polytechnical University\\
{\tt\small xiangmochu@mail.nwpu.edu.cn}
\and
Jing Zhang\\
Australian National University\\
{\tt\small zjnwpu@gmail.com}
\and
Nick Barnes\\
Australian National University\\
{\tt\small nick.barnes@anu.edu.au}
\and
Yuchao Dai\\
Northwestern Polytechnical University\\
{\tt\small daiyuchao@gmail.com}
}
\maketitle

\begin{abstract}
Effectively measuring and modeling the reliability of a trained model is essential to the real-world deployment of monocular depth estimation (MDE) models. However, the intrinsic ill-posedness and ordinal-sensitive nature of MDE pose major challenges to the estimation of uncertainty degree of the trained models.
On the one hand, utilizing current uncertainty modeling methods may increase memory consumption and are usually time-consuming.
On the other hand, measuring the uncertainty based on model accuracy can also be problematic,
where uncertainty reliability and prediction accuracy are not well decoupled.
In this paper, we propose to model the uncertainty of MDE models from the perspective of the inherent probability distributions originating from the depth probability volume and its extensions, and to assess it more fairly with more comprehensive metrics.
By simply introducing additional training regularization terms, our model, with surprisingly simple formations and without requiring extra modules or multiple inferences, can provide uncertainty estimations with state-of-the-art reliability, and can be further improved when combined with ensemble or sampling methods.
A series of experiments demonstrate the effectiveness of our methods.


\end{abstract}

\section{Introduction}
\label{sec:intro}

Monocular depth estimation (MDE) aims to estimate the depth of a scene from a single RGB image. Based on the depth representation, conventional MDE methods can be roughly grouped into regression approaches~\cite{ranftl2021vision, alhashim2018high, yin2021learning} and classification approaches~\cite{Bhat_2021_CVPR, li2018monocular, fu2018deep}.
Although significant progress~\cite{ranftl2021vision, li2022depthformer, li2022binsformer} focusing on improving model accuracy has been made, especially with transformer architectures~\cite{dosovitskiy_ViT_ICLR_2021, liu2021swin}, we find that the lack of model reliability indicators poses challenges for real-life deployment of the MDE models.
For example, an over-confident MDE model within an autonomous driving system will cause severe damage, and a better understanding of the model predictions can then avoid such disasters with better decision making.
For safe real-world deployment, we argue that both uncertainty estimation~\cite{kendall2017uncertainties} methods to explain model predictions, and uncertainty estimation measures to evaluate the uncertainty estimation techniques are desirable for MDE.


A systematic way to deal with uncertainty is via Bayesian statistics.  
Bayesian Neural Networks (BNNs) \cite{Bayesian_Uncertainty,bayesian_conv, generalized_bnn, deup_arxiv,canyoutrust,modern_bnn,kendall2017uncertainties,uncertainty_decomposition} aim to learn a distribution over each of the network parameters by placing a prior probability distribution over network weights. Calculating the exact Bayesian posterior is computationally intractable. 
Many approaches have aimed to develop approximations of BNNs that can work in practice, \ie~Monte Carlo Dropout~\cite{mcdropout}. For monocular depth estimation~\cite{ming2021deep}, two main streams exist for uncertainty estimation: 1) self-supervised methods~\cite{poggi2020monouncertainty,hirose2021variational,nie2021uncertainty,zhou2021sub,choi2021adaptive} usually obtain uncertainty following the noise-corruption model from~\cite{kendall2015bayesian}, where the estimated uncertainty serves as both weight of the reconstruction loss function and regularizer term to prevent the learned uncertainty from dominating the training process; 2) MDE models either design an auxiliary uncertainty estimation head~\cite{latent_Discriminant_deterministic_uncertainty} to regress prediction error or directly compute uncertainty with the post-hoc techniques~\cite{Hornauer2022GradientbasedUF}.


%



Due to the close correlation between uncertainty estimation and model calibration~\cite{Calibration_guo_2017_pmlr},
uncertainty is usually evaluated with modal calibration measures, \ie expected calibration error~\cite{degroot1983comparison}, negative log likelihood.
In addition to these measures, for depth estimation and optical flow estimation, \textit{sparsification errors} and \textit{sparsification plots}~\cite{ilg2018uncertainty} are used to represent model calibration degree. Area under sparsification curve is widely used as uncertainty measure for self-supervised/monocular depth estimation~\cite{poggi2020monouncertainty,Hornauer2022GradientbasedUF}.




Although reasonable uncertainty can be generated, we find that ignorance of the \enquote{ordinal-aware} 
attribute of MDE is one of the critical issues of existing uncertainty methods for MDE. 
Depth is an ordinal measure, and mis-classifications can lead to errors with different degrees of severity depending on the \enquote{relative} distance between labels.
Further, we observe that \textit{sparsification plots}~\cite{ilg2018uncertainty} and their variants, \ie~area under sparsification curve, cannot produce reliable uncertainty measures. A decrease in sparsification errors does not necessarily originate from improved uncertainty reliability, and it could also come from improved accuracy, making it not suitable to evaluate the uncertainty quality of different models.

We 
contribute to
reliable uncertainty estimation and evaluation for monocular depth estimation.
For the former, we first introduce a simple and effective \enquote{train-time} deterministic uncertainty generation method considering the ordinal-aware attribute of our task; then, inspired by deep metric learning~\cite{chopra2005learning,weinberger2005distance,sohn2016npair}, we incorporate a proper regularization, namely ranking loss~\cite{moon2020confidence} between error and uncertainty, achieving uncertainty-aware learning. For the latter, we propose to measure the uncertainty degree of monocular depth estimation from the inherent probability distribution.
Specifically, we adopt a Spearman correlation coefficient to provide an intuition of how faithfully a model can provide monotonic relationships between error and uncertainty,
which is 
demonstrated to be
more suitable to evaluate the generated uncertainty across different models. 
We also provide expressive uncertainty visualizations to better demonstrate the effectiveness of our method.


Our main contributions are summarized as: 
\begin{enumerate}[1)]
    \item We introduce an ordinal-aware train-time uncertainty generation method, which can be used during both training and testing for deterministic uncertainty generation.
    \item We present effective regularizer
    in training without relying on auxiliary networks to perform ordinal-aware and uncertainty-aware training, which is 
    demonstrated to be
    effective in generating high-quality uncertainty.
    \item We show that 
    standalone
    sparsification errors/plots 
    are
    less effective in evaluating uncertainty quality of
    different models, and then adopt a Spearman correlation coefficient to measure the uncertainty quality, providing monotonic relationships between error and uncertainty.
    \item {We propose to visualize model uncertainty from the perspective of depth probability volume, utilizing volume rendering techniques.}
\end{enumerate}

\section{Related Work}
\label{sec:related_work}

\subsection{Monocular Depth Estimation Models}
According to the types of supervision, existing MDE methods can be roughly grouped into supervised methods, self-supervised methods and weakly-supervised methods.
Supervised methods~\cite{saxena2007learning, eigen2014depth, ranftl2020towards} learn the image-to-depth mapping directly with the ground truth depth map as supervision.
Self-supervised methods learn depth from geometric consistency from stereo image pairs~\cite{garg2016unsupervised, Gonzalez_2021_CVPR} or consecutive frames~\cite{zhou2017unsupervised, godard2019digging}. To ease the label generating process, weakly-supervised methods~\cite{chen2016single, ren2020suw} learn depth from ordinal annotations.
With regard to depth scales, the metric methods~\cite{yin2021learning, fu2018deep, li2018monocular, alhashim2018high, Bhat_2021_CVPR} provide depth with physical scales and the metric-free methods~\cite{eigen2014depth, eigen2015predicting,  ranftl2020towards, ranftl2021vision} produce relative depth relations within an image. Based on the different depth representations,
regression methods~\cite{ranftl2021vision, alhashim2018high, yin2021learning} directly regresses a one-channel depth map or inverse-depth map from the RGB image. The classification methods~\cite{Bhat_2021_CVPR, li2018monocular, fu2018deep} model MDE as a classification task to obtain the probability distribution over depth scales. The final depth map is generated via soft weighted sum or maximum probability selection.
In this work, we focus on supervised MDE for metric depth estimation.

\subsection{Uncertainty Estimation}

Existing uncertainty estimation techniques can be roughly divided to ensemble solutions~\cite{mcdropout,deepensemble}, auxiliary uncertainty predictions~\cite{kendall2017uncertainties,latent_Discriminant_deterministic_uncertainty,postels2020hidden,latent_Discriminant_deterministic_uncertainty}, deterministic test-time uncertainty generation methods~\cite{hendrycks17baseline_softmaxconf,gradients_detect_distribution_shift,mi2022training} 
\etc.
The ensemble of predictions can be achieved by
introducing randomness in model inference~\cite{mcdropout}, or by designing an ensemble structure explicitly~\cite{deepensemble}.
Bayesian SegNet~\cite{kendall2015bayesian} adopts Monte Carlo Dropout~\cite{mcdropout} for semantic segmentation.
Infer-perturbations~\cite{mi2022training} adds random noise to the deep feature during inference, achieving training-free uncertainty in image super-resolution and depth estimation. 
The auxiliary uncertainty head can be introduced to model uncertainty based on prior knowledge.
Kendall \etal ~\cite{kendall2017uncertainties} and Asai \etal \cite{asai2019multi} use an auxiliary network to predict data uncertainty, which is jointly trained with the main task, \ie~semantic segmentation.
SLURP~\cite{yu2021slurp} and Hu \etal~\cite{hu2021learning} learn the error of the output using a second network, which will be used to produce label-free uncertainty at test time.

\begin{figure*}[tb]
\centering
\includegraphics[width=\linewidth, trim=0 0 0 0, clip]{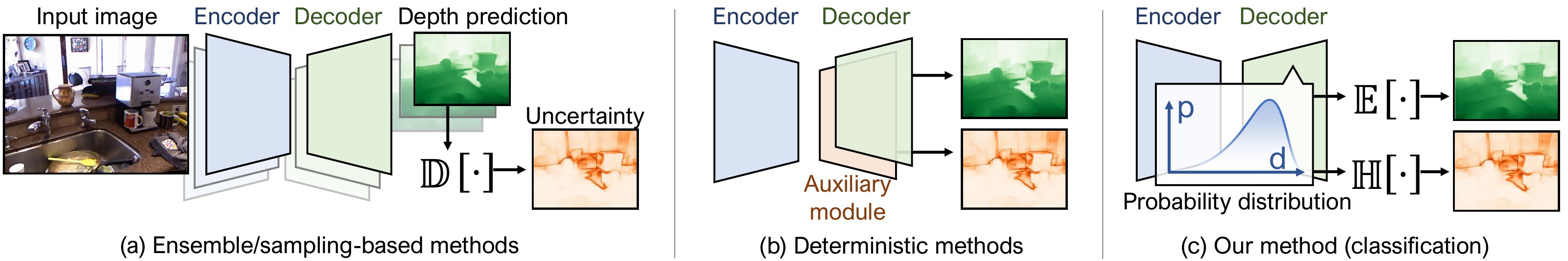}
\vspace{1pt}
\caption{A comparison of uncertainty derivation of different methods. $\mathbb D[\cdot]$, $\mathbb E[\cdot]$, $\mathbb H[\cdot]$ mean variance, numerical expectation and entropy. a) shows that ensemble~\cite{deepensemble} and sampling~\cite{mcdropout} based methods require multiple forwards or mapping functions to provide uncertainty to generate
multiple predictions; b) indicates that deterministic methods~\cite{postels2020hidden,latent_Discriminant_deterministic_uncertainty,addressing_failure_prediction_modeling_confidence} usually require extra network components to estimate uncertainty; c) illustrates the principle of our methods, where the uncertainty is directly generated from the probability distribution.}
\label{fig:uncert_comparison}
\end{figure*}

For MDE, \cite{Hornauer2022GradientbasedUF} introduces gradient based confidence estimation, where the uncertainty is obtained as the gradient of a feature map, given an auxiliary loss.
\cite{latent_Discriminant_deterministic_uncertainty} learns a set of prototypes, and uses distinction maximization~\cite{macedo2021entropic} to connect sample distance with prototypes, where uncertainty is produced with an auxiliary uncertainty estimation head. For self-supervised depth estimation, \cite{hirose2021variational} introduces the Mahalanobis-Wasserstein distance between two consecutive frames to learn the uncertainty. \cite{poggi2020monouncertainty} carries out extensive studies to learn uncertainty following the image-reconstruction pipeline. \cite{zhou2021sub} estimates the confidence of the LiDAR sparse depth map to filter out outliers for robust training. Built upon~\cite{kendall2017uncertainties}, \cite{nie2021uncertainty,choi2021adaptive} learn uncertainty based on the noise-corruption model. 
\cite{hu2022uncertainty}decomposes predictive uncertainty as error variance (caused by inherent noise) and estimation variance (caused by limited training data); a conditional CDF is constructed to achieve ordinal-aware learning, which is interesting and 
has been demonstrated to be effective.
A comparison between different uncertainty modeling methods is shown in Fig.~\ref{fig:uncert_comparison}.

\subsection{Uncertainty Measures}
\label{sec:uncertainty measures}
Uncertainty can be measured indirectly based on the calibration degree of the related model~\cite{Calibration_guo_2017_pmlr}, where the assumption is that uncertainty of a well-calibrated model should be consistent with prediction error.
The widely studied calibration measures include expected calibration error \cite{degroot1983comparison}, negative log-likelihood, entropy of prediction, 
\etc.

\textit{Expected Calibration Error (ECE)}~\cite{degroot1983comparison} is a bin-based strategy to measure the expectation difference between model confidence and model accuracy.
\textit{Negative log likelihood (NLL)} for a model $f_\theta(y|x)$ is defined as $\text{NLL} = -\sum_{i=1}^N \log \left(f_\theta(y_i|x_i)\right)$, which is minimized if and only if $f_\theta(y_i|x_i)$ recovers the true conditional distribution of $f(y|x)$ \cite{Calibration_guo_2017_pmlr}. \textit{Entropy} is a convenient way to model the state of disorder/randomness or uncertainty, which is directly achieved by computing entropy of model prediction.

Different from classification-based calibration measures, \ie,~ECE~\cite{degroot1983comparison}, calibration error for regression measures the expected confidence interval, \eg, the prediction should fall into the 90\% confidence interval 90\% of the time. Expected normalized calibration error (ENCE)~\cite{levi2022evaluating} extends ECE to measure the calibration degree of regression models.
\cite{cui2020calibrated} proposes to use maximum mean discrepancy (MMD) to perform distribution matching between the regression ground truth target and the random samples from the predictive distribution. ~\cite{kuleshov2018accurate} demonstrates training an auxiliary model on top of a pre-trained forecaster, and the experiment on MDE shows a closer expected confidence level with the observed one. ~\cite{song2019distribution} focus on obtaining well-calibrated output distributions from regression models with a post-hoc calibration method. Similarly, ~\cite{zelikman2020crude, levi2022evaluating, thiagarajan2020building, zhao2020individual} propose various methods to perform model calibration for regressors.

In addition to
the indirect calibration measures, uncertainty can also be measured directly with uncertainty measures, \ie~sparsification errors/plots~\cite{ilg2018uncertainty} and its variants.
For self-supervised MDE~\cite{poggi2020monouncertainty, yu2021slurp, latent_Discriminant_deterministic_uncertainty} and optical flow estimation~\cite{bruhn2006confidence, kondermann2008statistical, kybic2011bootstrap, marquez2012complete, mac2012learning, wannenwetsch2017probflow}, the \textit{Area Under Sparsification Curve (AUSE)} is usually used to measure the misalignment between the confidence and the accuracy. However, it does 
not
reveal how strongly these two measurements are related, \cite{wannenwetsch2017probflow} use Spearman’s rank correlation coefficients~\cite{myers2004spearman} to model how well the uncertainty can be mapped into the error using an arbitrary monotonic function.

\subsection{Uncertainty from Probability Volume}
\label{sec:probability volume}

A probability volume is a data structure containing the estimated probabilities \wrt discretized hypotheses, targeting learning correspondence from image pairs. It is widely used in tasks like stereo matching \cite{chang2018pyramid, kendall2017end, zhang2020adaptive, shen2021cfnet}, multi-view stereo~\cite{huang2018deepmvs, yan2020dense, ding2022transmvsnet, peng2022rethinking} and optical flow estimations~\cite{yang2019volumetric, wang2020displacement}. Usually, \textit{cost volumes} are constructed to store the cost for matching hypotheses.

In comparison, classification-based MDEs learn the probabilities of the depth hypothesis of every pixel in an image, so the feature of the penultimate layer is actually a depth probability volumes~\cite{liu2019neural}.
Though in this situation, rather than denoting correlations between hypotheses feature pairs, the values in the probability volumes are purely from a single image. Extracting uncertainty from probability volumes is an effective tool that can improve the interpretability of a model. \cite{liu2019neural} visualizes confidence from the depth probability volume of a moving camera to verify the model's effectiveness to potentially boost downstream tasks. \cite{wang2020displacement} explicitly involves entropy of the matching cost in feature updating for optical flow estimation. 

Unlike other dense prediction tasks, \ie,~semantic segmentation~\cite{kendall2017uncertainties}, MDE is ill-posed and ordinal-sensitive, making reliable uncertainty estimation especially necessary. For the first time, we introduce an uncertainty estimation method and measure to explore the ordinal-sensitive nature of MDE with 3D visualization of the depth probability volume. 
Our ordinal-sensitive uncertainty estimation model, uncertainty measure, and our solution of visualizing the uncertainty of MDE using 3D depth probability volume make our work significantly different from existing techniques.

\section{Method}
We first introduce classification-based MDE models, discuss its \enquote{ordinal-sensitive} nature and present our interpretation of depth prediction as a depth probability volume in Sec.~\ref{subsec_mde_ordinal_aware}. Then we derive the deterministic uncertainty in Sec.~\ref{subsec_uncertainty_derivation}, based on which a ranking-based regularization is proposed in Sec.~\ref{subsec_ranking_loss} to model the ordinal-sensitive nature of MDE. We present our new uncertainty measure in Sec.~\ref{sec:sparsification flaw}, and introduce the 3D depth probability volume as an uncertainty visualization tool
in Sec.
\ref{sec:discussion}.

\subsection{MDE as Ordinal-aware Classification}
\label{subsec_mde_ordinal_aware}

We define the training dataset as $\mathcal D=\left\{x_i, d_i\right\}_{i=1}^{N}$, consisting of $N$ pairs of RGB images $x_i$ and a one-channel depth map $d_i$ of shape $H\times W$, where $i$ indexes the samples.
The regression-based MDE models directly regress a one-channel depth map $\hat{d}=f_\theta(x)$, where $\theta$ represents the network parameters to achieve the image-to-depth mapping.

Though being a dense regression task, MDE can be solved in a classification fashion. 
Continuous depth values are discretized into $M$ increasing depth hypotheses $\mathbf s=\left[s_1, s_2, ..., s_M\right]^\mathrm{T}\in\mathbb{R}^{M}$, which can be pre-defined uniformly in linear space or log space~\cite{DORN_fu_2018_cvpr}, or can be predicted by a separate network~\cite{Bhat_2021_CVPR}. 
{For every pixel in the image,} the network is trained to predict probabilities $\mathbf p^{(h,w)}=[p_1, p_2, ..., p_M]^\mathrm{T}\in\mathbb{R}^{M}$  (where $(h,w)$ is the pixel coordinate), denoting the depth value for a certain pixel falling into these intervals. 
For evaluation, the numerical expectation of such a probability distribution $\hat{d}=\mathbf s^\mathrm{T}\cdot \mathbf p$ ($\cdot$ represents the inner product
) is used to represent the estimated depth.

\noindent\textbf{Depth prediction reinterpretation:} We can interpret the predicted probability distributions in a geometric way: each estimated probability located at $(h,w,m)$ represents the possibility that there exists 
a
physical 3D point with depth $s_m$, being projected to the 2D plane at $(h,w)$, \ie, the location of probabilities agree with the physical points within the viewing frustum, so the predicted probabilities act as a depth probability volume.

\noindent\textbf{The ordinal-sensitive issue:} Using a classification approach to achieve depth regression 
has been demonstrated
to be beneficial for fine-grained depth confidence estimation~\cite{Bhat_2021_CVPR}.
Further,
the estimated depth probability distribution
contains richer information than the regressed
depth value itself. There are not only the probabilities that a point belongs to \textit{each} of the depth hypothesis intervals, but also the hidden uncertainty of the prediction~\cite{shen2021cfnet}.
However, such a property is not innate for classification-based MDEs, 
since supervision that only uses regression is not strong enough to help the network distinguish the relative ordinal relationships between different depth hypotheses.
This motivates our ordinal-sensitive uncertainty estimation/measure
to be consistent with the ordinal-sensitive nature of MDE. 

{To understand the representation advantage of classification-based MDE networks, consider that they produce probabilities for the depth falling into intervals. So when the network is certain about a prediction, it should produce a unimodal distribution that is tightly clustered. Alternatively, uncertain predictions can result in network predictions of multiple possible locations where an object can occur, resulting in a spread-out distribution.
}

Regression models, however, do not benefit from this geometric interpretation because they 
{produce the final regressed value from latent variables and} lack such probabilistic representation.
The demonstrated experiments show that by imposing effective regularization terms, we can still extract uncertainty with similar approaches.


\subsection{Uncertainty Derivation}
\label{subsec_uncertainty_derivation}

\begin{figure}[t]
\begin{center}
    {\includegraphics[width=\linewidth]{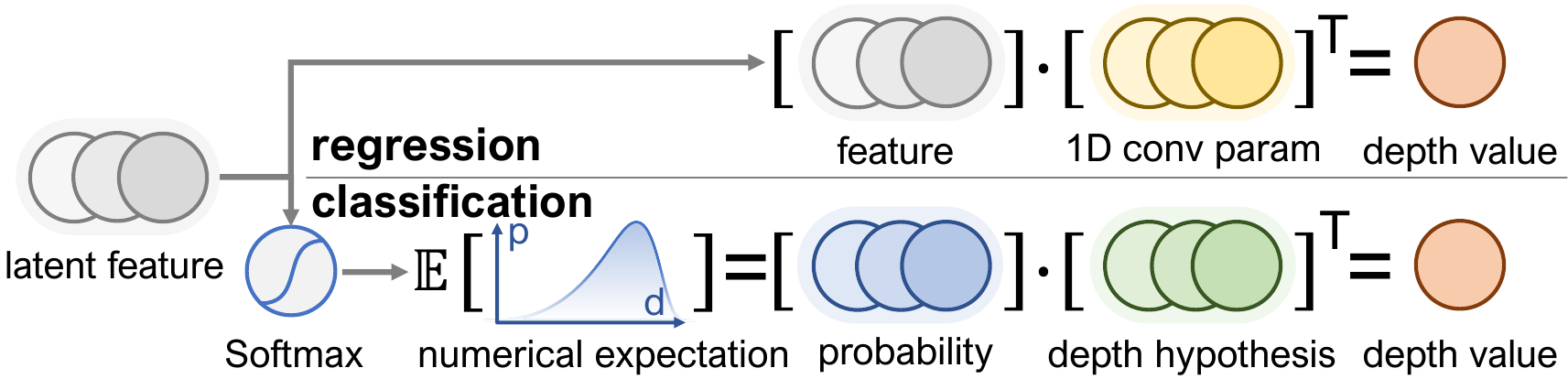}}
\end{center}
\caption{A comparison between regressors of classification based and regression based MDE methods.
}
\label{fig:unified_cls_reg}
\end{figure}

As a dense prediction task, MDE networks usually consist of an encoder, a decoder and a regressor.
The encoder, or the backbone network, is usually pretrained on image classification tasks and can provide features in different sizes.
The decoder gradually upsamples and fuses the features from the encoder network.
The regressor generates the prediction as a one-channel depth map.
In Fig.~\ref{fig:unified_cls_reg}, we show how regression and classification based MDEs work: a network maps an input image $x$ to a latent feature map $\mathbf z$ as the penultimate feature in the regressor, then the difference between the two types of methods (regression MDE and classification MDE) takes place.


We introduce a random variable $D$ to represent the depth prediction.
For classification based MDEs, we can obtain a probability distribution $\mathbf p$ by applying the Softmax operation along the feature dimension, and the final depth values are the numerical expectations of
$D$:\footnote{We omit superscript $(h,w)$ for simplicity if operations are applied equally to any position in the map.}
\begin{equation}
    \hat d = \mathbb{E}(D) = \mathbf s^\mathrm{T}\cdot \mathbf p = \mathbf s^\mathrm{T} \cdot\text{Softmax}(\mathbf z).
\end{equation}

Since the estimated probability is an inherent source of uncertainty~\cite{hendrycks17baseline_softmaxconf}, we can measure its entropy and regard the entropy as an indicator of uncertainty via:
\begin{equation}
\label{ue_defination}
    u = \alpha \cdot \mathbb{H}(D) = -\alpha \cdot \sum_{m=1}^M p_m \cdot \log p_m,
\end{equation}
where the scalar $\alpha=\text{SoftPlus}(a)$ is a learnable parameter used to adjust the numerical range of the uncertainty for more stable training. Finally, we get the uncertainty map $u$ of shape $H\times W$.

\noindent\textbf{Uncertainty for regression-based MDE:} For regression-based MDE, usually a convolutional layer is applied on $\mathbf z$ to produce the final one-channel depth map via: $\hat d=\mathbf w^\mathrm{T} \cdot \mathbf z$ (see \enquote{regression} branch of Fig.~\ref{fig:unified_cls_reg} with a 1$\times$1 convolutional layer):
Though such representation does not involve probability, we can still measure the entropy of a pseudo probability $\text{Softmax}(\mathbf z)$ and get the uncertainty via Eq.~\ref{ue_defination}. 
The intuition of doing so is that since regression and classification methods 
have
similar behaviors, introducing extra constraints on the pseudo probability can bring shared characteristics of indicating uncertainties.
We show in the experimental results section that by imposing extra regularization on this term, we can get uncertainty estimations with decent degrees of correlations with the error.


\subsection{Uncertainty-aware Training via Ranking Loss}
\label{subsec_ranking_loss}

With the deduced uncertainty in Eq.~\ref{ue_defination}, the classification and regression based models can both provide uncertainty estimations.
To directly optimize the model for uncertainty-aware learning, we further introduce a ranking-based regularizer to link model uncertainty to prediction error.

We design the training loss with three components.
Firstly, we evaluate the depth prediction $\hat d$ using an $L_1$ loss:
\begin{equation}
    \mathcal L_r = \sum_{(h,w)}|\hat d^{(h,w)} - d^{(h,w)}|.
\end{equation}

Then, for classification methods, we add a soft label loss to regularize the probability distribution:
\begin{equation}
    \mathcal L_p = \sum_{(h,w), m} | y^{(h,w)}_m - p^{(h,w)}_m|,
\end{equation}
where $\mathbf y=\left[y_1, y_2, ..., y_M\right]\in\mathbb{R}^{M}$ is the soft label~\cite{SoftLabel_diaz_2019_cvpr} with 

\begin{align}
    y_m = \frac{e^{-\phi(s_m, d)}}{\sum_{\tilde{m}=1}^M e^{-\phi(s_{\tilde{m}}, d)}}, 
\end{align}
here $\phi(s_m, d)= \gamma \left|s_m - d \right|$; $\gamma=20$ is the hyper-parameter adjusting the shape of the soft label.
We add this term to apply constraints on the shape of the probabilities, since we expect confidence estimation to have a probability distribution with high concentration around the ground truth,
and with dispersed distribution to express uncertainty.
This term is 
set to
zero for regression methods, since the pseudo probability has no ordinal labels.

Thirdly, to regularize the uncertainty estimation, we add a ranking loss ~\cite{moon2020confidence} between error and uncertainty to model the ordering relationship between two quantities, inspired by 
deep metric learning~\cite{chopra2005learning,weinberger2005distance,sohn2016npair} via:

\begin{equation}
\label{loss_reg}
\begin{aligned}
    \mathcal L_u = \sum_{(h,w)} \max \Big \{ 0, &\left[r ^{(h,w)}-r^{(h',w')}\right]_\text{sg} \\
     -&\left[u^{(h,w)} - u^{(h',w')}\right] \Big\},
\end{aligned}
\end{equation}
where $r=|\hat d - d|$ is the error, 
$u$ is the uncertainty estimation in Eq.~\ref{ue_defination}.
The stop gradient operator $\left[\cdot\right]_\text{sg}$ detaches the gradient of the variable, so that the term will not be involved in the back-propagation. Here, we encourage the ranking of the uncertainty to match that of the error, but do not want the error term to adapt to the estimated uncertainty.
We can randomly choose another location $(h',w')$ from the prediction to provide $r^{(h',w')}$ and $u^{(h',w')}$ following the contrastive learning pipeline~\cite{sohn2016npair}.
In practice, we randomly shuffle the uncertainty map and the depth prediction synchronously to get $r'$ and $u'$, and measure the pixel-wise truncated disagreement between the difference in error and the difference in uncertainty.

The final loss is composed of $\mathcal L_r$, $\mathcal L_p$ and $\mathcal L_u$, and they are automatically weighted following~\cite{kendall2018multi} so that we do not need to tune their weights.
The final loss takes the form as:
\begin{equation}
    \mathcal L = \sum_{i\in\{r,p,u\}}\mathcal L_{i}/\exp(\sigma_i)+\sigma_i,
\end{equation}
where $\sigma_r$, $\sigma_p$ and $\sigma_u$ are learned along with the model parameters, which serve as both weight for the loss term and regularizer~\cite{kendall2017uncertainties}.

\noindent\textbf{Extension to regression based MDE:} We show in our experiments that with extra terms of regularization $\mathcal L_u$, the uncertainty from the probability distribution is comparable to the state-of-the-art monocular depth uncertainty methods, with high correlations to the prediction error.
Further, we can extend such a method to regression based MDE by measuring the entropy ($u$ in Eq.~\ref{ue_defination}) from the unordered pseudo probabilities and imposing a similar regularization term.

\subsection{Uncertainty Measure with Spearman Correlation Coefficients}
 \label{sec:sparsification flaw}
 
\noindent\textbf{Is sparsification error good enough to measure model uncertainty?}
Sparsification plots reveal how much the estimated uncertainty coincides with the factual errors~\cite{ilg2018uncertainty}, which is obtained via measuring model accuracy by gradually removing the highest value from uncertainty/error, obtaining both sparsification and the oracle, respectively.
Sparsification errors \cite{mac2012learning, ilg2018uncertainty, poggi2020monouncertainty} are used to assess the uncertainty in previous works, which is defined as a gap between sparsification and the oracle, and the area between them is the Area Under Sparsification Error (AUSE). Depending on the different error metrics 
that are used
, sparsification errors are named after a specific error measure.

This is reasonable since a good uncertainty map should look similar to the error map, leading to a small sparsification error and a small AUSE.
However, the error itself usually has its own metric, and the sparsification error also has its metric. This works fine when comparing methods with the same degree of accuracy, but sparsification errors between models with different accuracy degrees cannot accurately tell the uncertainty quality. In other words, with the same degree of correlation between uncertainty and error, a model with higher accuracy can have lower sparsification error, which can not reveal the true uncertainty quality.

\noindent\textbf{Spearman correlation coefficients for uncertainty measure:} We claim that an ideal uncertainty should indicate where the model would make mistakes, and uncertainty measurements should reflect how correlative the uncertainty is with the actual error, rather than 
reflecting
the error itself. 
We find the straightforward measurements, \ie,~Spearman correlation coefficients~\cite{myers2004spearman}, are comparable across models with different degrees of accuracy, and are used in~\cite{mi2022training, asai2019multi, wannenwetsch2017probflow}. 

Spearman correlation coefficients are measured on the ranking of two variables, which can provide evidence of how strongly the monotonic relationships between the two sets of values, regardless of their magnitudes or making assumptions about their underlying relationships, \eg, linear, quadratic, \etc. In this case, we use Spearman correlation coefficients as an alternative uncertainty measure to evaluate the monotonic relationship between model uncertainty and prediction error. 
This measurement directly validates the assumption that the higher the uncertainty, the higher the error, which is consistent with the criteria of model calibration~\cite{Calibration_guo_2017_pmlr}.
The fact that some recent work~\cite{latent_Discriminant_deterministic_uncertainty, yu2021slurp, hu2022uncertainty, poggi2020monouncertainty} only adopt sparsification errors for uncertainty evaluation triggered our concern to assess uncertainty more fairly.



\section{Experimental Results}

\subsection{Setup}
\label{subsec:setup}
\noindent\textbf{Dataset:} We evaluate our methods on two commonly used MDE datasets, namely 
the
NYU Depth V2~\cite{silberman2012indoor} and KITTI~\cite{geiger2013vision} datasets. For NYU, we use the pre-processed data provided by~\cite{alhashim2018high} with 50K training samples and 654 testing samples following \cite{eigen2014depth}. For the KITTI dataset, we use a training dataset with 26K images from the left view, and a testing dataset with 697 images as specified in~\cite{eigen2014depth, Bhat_2021_CVPR}.

\noindent\textbf{Implementation Details:}
We evaluate the proposed measures on various models using the two datasets. 
Linear discretization of depth value is adopted for simplicity. 
Each model is trained for 10 epochs with an initial learning rate of 1e-4 and decay of 0.8 every 2 epochs. 
Training and testing is
on a single NVIDIA GeForce RTX 3090 GPU. 
Implenentation details
can be found in the supplementary material. 

\subsection{Measures}

\noindent\textbf{Evaluation Metrics:} 
We use three metrics to evaluate model accuracy, including
RMSE: $\sqrt{\frac 1 N \sum_{i=1}^N (d_i - \hat d_i)^2}$,
Rel: $\frac 1 N \sum_{i=1}^N \frac{\left|d_i - \hat d_i\right|} {d_i}$,
and $\delta_1$: $\%$of $d_i$ s.t. $\max \left(\frac {d_i} {\hat d_i}, \frac {\hat d_i} {d_i}\right)=\delta < 1.25$. Since $\delta_1$ thresholds the error map and averages the binary values, following~\cite{qu2022improving, yu2021slurp}, we adopt the AUROC metric to represent the separability of the uncertainty of predictions that lie inside and outside the threshold. 

To measure uncertainty, we adopt sparsification errors~\cite{ilg2018uncertainty} following conventional practice~\cite{mac2012learning,ilg2018uncertainty,poggi2020monouncertainty}, 
where AUSE is used to measure the difference between the estimated and oracle sparcification. As sparsification is defined based on a given error metric, different error metrics will lead to different AUSE.
Further, we use a Spearman correlation coefficient (SCC in Table~\ref{Tab:backbone_nyu} and~\ref{Tab:backbone_kitti}) to provide an intuition of how faithfully a model can provide monotonic relationships between error and uncertainty.
The coefficients are measured in image level, between pixel-wise $L_1$ error and uncertainty, and are averaged over the testing dataset.

\begin{figure*}[htbp]
\centering
\includegraphics[width=\linewidth]{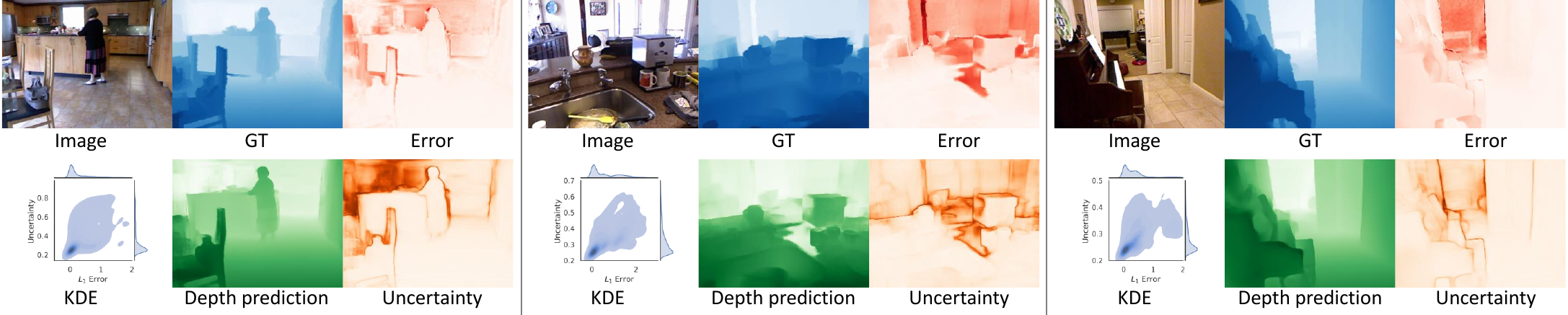}
\caption{\small{Visual results on NYU, estimated by the classification based model adopting Swin transformer as the backbone (Ours-C,S).
We show the Kernel Density Estimation (KDE)~\cite{parzen1962kde} plot of the $L_1$ error and the uncertainty to better show the correlation between them.}}
\label{fig:visual}
\end{figure*}

\subsection{Results}
We first compare state-of-the-art uncertainty methods with ours in Table~\ref{Tab:backbone_nyu}~and \ref{Tab:backbone_kitti}. We list the AUSE of a certain error metric on its left-hand side. We denote the backbone of each method in the column \enquote{B}, where \enquote{R} means ResNet50~\cite{he2016deep_resnet}, \enquote{D} means DenseNet161~\cite{huang2017densely}, and \enquote{S} means Swin Large~\cite{liu2021swin} (we used the implementation of~\cite{yuan2022newcrfs} to support  unconstrained input image size). The decoders of our models are the same as~\cite{godard2019digging}. \enquote{Ours-C} and \enquote{Ours-R} represent the classification and regression-based methods, respectively.

The depth network architectures of LDU~\cite{latent_Discriminant_deterministic_uncertainty} and SLU (short for SLURP)~\cite{yu2021slurp} are BTS~\cite{lee2019big}, we follow their original implementation to train and evaluate on the two datasets.
Deep Ensemble (DE)~\cite{deepensemble}, MC Dropout (MCD)~\cite{mcdropout} and Infer-perturbations (Noise)~\cite{mi2022training} adopt the same architecture as \enquote{Ours-C}, where we attached three decoders and regressors to achieve DE~\cite{deepensemble}.
For MCD~\cite{mcdropout}, we added dropout operations to every skip connection. During inference, we forward MCD~\cite{mcdropout}, and Infer-perturbations~\cite{mi2022training} for 3 iterations, and measure the variance of the predictions as the uncertainty. 
Note that due to different measuring methods, readers may find inconsistencies between the comparing results of \cite{yu2021slurp, latent_Discriminant_deterministic_uncertainty} in Table~\ref{Tab:backbone_nyu}, Table~\ref{Tab:backbone_kitti} and the metrics reported in their original papers. We train these methods locally based on their open-sourced code, and measure the sparsification error with the implementation of~\cite{poggi2020monouncertainty} on all methods for fair comparisons.

\begin{table}[ht!]
\centering
\footnotesize
\renewcommand{\arraystretch}{1.2}
\renewcommand{\tabcolsep}{0.1mm}
\caption{ Accuracy and uncertainty comparisons on NYU. }
\resizebox{\linewidth}{!}{
 \begin{tabular}{r|c|cc|cc|cc|c}\toprule
     Method~ & ~B~ & ~RMSE$\downarrow$~ & ~AUSE$\downarrow$~ & ~~~Rel$\downarrow$~~~ & ~AUSE$\downarrow$~ & ~~~$\delta_1\uparrow$~~~ & \scriptsize{AUROC}$\uparrow$& ~~SCC$\uparrow$~~ \\\midrule
     DE~\cite{deepensemble} & R & 0.6936 & 0.3165 & 0.2051 &  0.0954 & 0.6735 & 0.5933 & 0.1914 \\
     MCD~\cite{gal2016dropout} & R & 0.7054 & 0.3223 & 0.2126 & 0.1009 & 0.6660 & 0.5823 & 0.2013 \\
     Noise~\cite{mi2022training} & R & 0.6814 & 0.2877 & 0.1978 & 0.0875 & 0.6827 & 0.6186 & 0.2399 \\
     LDU~\cite{latent_Discriminant_deterministic_uncertainty} & D & \rs0.4015 & \rt{0.1257} & 0.1132 & 0.0635 & 0.8739 & 0.5621 & 0.3117 \\
     SLU~\cite{yu2021slurp} & D & \rf0.3970 & \rf0.1144 & \rf0.1100 & 0.0501 & \rf0.8845 & 0.6820 & \rf0.3406 \\ \midrule

Ours-C~ & R & 0.7280 & 0.2781 & 0.2132 & 0.0894 & 0.6553 & 0.6349 & 0.2525 \\
Ours-C~ & D & 0.4811 & 0.1534 & 0.1289 & 0.0504 & 0.8361 & 0.7217 & 0.3099 \\
Ours-C~ & S & 0.4275 & \rt{0.1257} & \rt{0.1124} & \rf0.0420 & \rs0.8813 & \rf0.7656 & \rs0.3284 \\ \hline

Ours-R~ & R & 0.7283 & 0.2767 & 0.2180 & 0.0911 & 0.6465 & 0.6257 & 0.2480 \\
Ours-R~ & D & 0.4787 & 0.1488 & 0.1310 & \rt0.0498 & 0.8347 & \rt0.7324 & 0.3124 \\
Ours-R~ & S & \rt{0.4196} & \rs{0.1189} & \rs{0.1118} & \rs0.0425 & \rt0.8794 & \rs0.7639 & \rt0.3281 \\
         \bottomrule
\end{tabular}}
\label{Tab:backbone_nyu}
\end{table}


\subsection{Performance Comparison}

\noindent\textbf{Performance comparison:} In Table~\ref{Tab:backbone_nyu} and~\ref{Tab:backbone_kitti}, we provide accuracy and uncertainty measurements of our and compared models. Our methods with different backbones provide varying performances, are comparable with the current state-of-the-art, but with significant advantages in memory consumption and computational complexity, \ie,~we do not involve extra networks or multiple forward sampling.

\noindent\textbf{Expressive uncertainty visualization:} 
Fig.~\ref{fig:uncert_vis}
shows the depth probability volume from our classification based model, where the network provides uncertain depth estimations for the contents in the mirror, which is consistent with the prediction error.

\noindent\textbf{Reliable uncertainty measure with Spearman correlation coefficients:} Fig.~\ref{fig:visual} shows the depth and the uncertainty,
along with the ground truth and error.
We can directly see the strong correlation between the uncertainty and error, and the high similarity between the estimated depth and the ground truth. The Kernel Density Estimation (KDE)~\cite{parzen1962kde} plot illustrates statistically more detailed relationships between error and uncertainty estimation.


\noindent\textbf{Speed and complexity comparison:} 
Table~\ref{tab:efficiency} compares run-time, memory and operational complexity of different models, which qualitatively shows the high efficiency of our method.

\begin{table}[t]
\centering
\footnotesize
\renewcommand{\arraystretch}{1.2}
\renewcommand{\tabcolsep}{0.1mm}
\setcounter {table} {1}
\caption{ \centering Accuracy and uncertainty comparisons on KITTI. }
\resizebox{\linewidth}{!}{
\begin{tabular}{r|c|cc|cc|cc|c}\toprule
     Method~ & ~B~ & ~RMSE$\downarrow$~ & ~AUSE$\downarrow$~ & ~~~Rel$\downarrow$~~~ & ~AUSE$\downarrow$~ & ~~~$\delta_1\uparrow$~~~ & \scriptsize{AUROC}$\uparrow$& ~~SCC$\uparrow$~~ \\\midrule

     DE~\cite{deepensemble} & R & 3.3819 & 0.7181 & 0.0946 & 0.0279 & 0.8994 & 0.8020 & 0.5419 \\
     MCD~\cite{gal2016dropout} & R & 3.3845 & 0.6574 & 0.1005 & 0.0295 & 0.8932 & 0.8083 & 0.5818\\
     Noise~\cite{mi2022training} & R & 3.3296 & 0.6025 & 0.0909 & 0.0227 & 0.9058 & 0.8453 & 0.5948\\
     LDU~\cite{latent_Discriminant_deterministic_uncertainty} & D & 3.0763 & 0.3256 & 0.0674 & 0.0200 & 0.9457 & 0.8433 & 0.6704\\
     SLU~\cite{yu2021slurp} & D &     2.9466 & 0.2662 &    0.0639 & 0.0140 &    0.9486 & 0.8948 & \rf0.7008\\ \midrule
Ours-C~ & R & 3.4090 & 0.3448 & 0.0970 & 0.0195 & 0.8952 & 0.8609 & \rt0.6862 \\
Ours-C~ & D & \rt2.5502 & \rt0.2352 & \rt0.0632 & 0.0139 & \rt0.9538 & 0.8985 & 0.6779 \\
Ours-C~ & S & \rs2.4095 & \rs0.2199 & \rf0.0571 & \rs0.0128 & \rs0.9651 & \rs0.9109 & 0.6735 \\ \hline

Ours-R~ & R & 3.4172 & 0.3408 & 0.0966 & 0.0193 & 0.8940 & 0.8657 & \rs0.6909 \\
Ours-R~ & D & 2.5716 & 0.2369 & 0.0641 & \rt0.0138 & 0.9529 & \rt0.8988 & 0.6778 \\
Ours-R~ & S & \rf2.3758 & \rf0.2072 & \rf0.0571 & \rf0.0127 & \rf0.9655 & \rf0.9125 & 0.6790 \\
    \bottomrule
\end{tabular}}
\label{Tab:backbone_kitti}
\end{table}

\begin{table}[b]
\setcounter{table}{2}
    \centering
    \caption{Model consumption on time, memory and operations. 
    All models adopt DenseNet161 as the backbone network and are measured under the same  condition.
    MCD and DE are with 3 forward passes and 3 heads respectively. The image size is 480$\times$640. 
    }
    \footnotesize
    \begin{tabular}{c|c c c c c} \toprule
        Method & SLU\cite{yu2021slurp} & LDU\cite{latent_Discriminant_deterministic_uncertainty} & MCD\cite{mcdropout} & DE\cite{deepensemble} & Ours \\ \midrule
        Time(ms) & 40.2 & 36.9 & 55.1 & 36.8 & \textbf{18.8}\\
        Params(M) & 87.2 & 47.0 & \textbf{31.1} & 40.3 & \textbf{31.1} \\
        MACs(G)  & 209.2 & 122.29 & 313.7 & 217.6 & \textbf{104.6}\\ \bottomrule
    \end{tabular}
    \label{tab:efficiency}
\end{table}

\subsection{Ablation Studies}

\begin{table}[t]
\centering
\footnotesize
\renewcommand{\arraystretch}{1.2}
\renewcommand{\tabcolsep}{0.1mm}
\setcounter{table}{3}
\caption{ The effect of removing loss terms, evaluated on KITTI, the base model is \enquote{Ours-C, R}, a classification model adopting ResNet as the backbone.}
\resizebox{\linewidth}{!}{
 \begin{tabular}{c c c |cc|cc|cc|c}\toprule
    $\mathcal L_r$ & $\mathcal L_p$ & $\mathcal L_u$ & ~RMSE$\downarrow$~ & ~AUSE$\downarrow$~ & ~~~Rel$\downarrow$~~~ & ~AUSE$\downarrow$~ & ~~~$\delta_1\uparrow$~~~ & \scriptsize{AUROC}$\uparrow$& ~~SCC$\uparrow$~~ \\\midrule
$\checkmark$ & $\checkmark$ & $\checkmark$ & \b{3.4090} & \b{0.3448} & \b{0.0970} & \b{0.0195} & \b{0.8952} & \b{0.8609} & \b{0.6862} \\ \midrule
$\checkmark$ & & & 3.4661 & 0.7473 & 0.0987 & 0.0302 & 0.8905 & 0.8062 & 0.4982 \\
$\checkmark$ & $\checkmark$ & & 3.4500 & 0.6229 & 0.0981 & 0.0271 & 0.8909 & 0.8159 & 0.5931 \\
$\checkmark$ & & $\checkmark$  & 3.4244 & 0.3503 & 0.0971 & 0.0208 & 0.8936 & 0.8523 & 0.6849 \\
$\checkmark$ & $\checkmark$ & $\blacklozenge$ & 3.4815 & 0.6037 & 0.0980 & 0.0305 & 0.8907 & 0.8086 & 0.5081 \\
$\checkmark$ & $\checkmark$ & $\blacksquare$  & 3.4325 & 0.4479 & 0.0965 & 0.0255 & 0.8942 & 0.8448 & 0.6114 \\
    \bottomrule
\end{tabular}}
\label{Tab:removing}
\end{table}

\noindent\textbf{The effect of loss terms:}
We verify the effect of our proposed loss terms by training the model \enquote{Ours-C,R} without using one of the proposed strategies.
Replacing the ranking loss in $\mathcal L_u$ with other terms could bring a significant decrease in uncertainty estimation: $\blacklozenge$ cancels the $\max(\cdot)$ operation in the ranking loss, $\blacksquare$ directly applies an $L_1$ loss between error and uncertainty.
With the results shown in Table~\ref{Tab:removing}, we can draw conclusions that \textbf{1)} with the joint effect of the three loss terms, the model can provide results with accurate predictions and reasonable uncertainty; \textbf{2)} classification models can inherently provide decent uncertainty; \textbf{3)} the proposed simple regularization term ($\mathcal{L}_u$) can further boost the performance of the raw model.

\subsection{Observations}


\noindent\textbf{Lower sparsification errors do not necessarily mean better uncertainty.}
Though being widely adopted, we argue that higher accuracy or better uncertainty could both lead to a decrease in sparsification errors, and sparsification errors from models with different accuracy are not comparable.
Our classification method with DenseNet backbone (denoted in \enquote{Ours-C, D}) has similar uncertainty with LDU~\cite{latent_Discriminant_deterministic_uncertainty} when assessing the Spearman correlation coefficient, while our model shows lower AUSE-RMSE when the model accuracy is better on KITTI, but higher AUSE-RMSE when the model accuracy is worse on NYU.
The conclusion still holds if we compare the same model trained and evaluated separately on the two datasets, where all models provide consistently better uncertainty on KITTI, but the AUSE-RMSE metric presents the opposite conclusion.
All these pieces of evidence indicate that the model accuracy is highly involved in sparsification metrics, making it not ideal to evaluate the uncertainty qualities of different models.

\noindent\textbf{Better backbones lead to better accuracy and uncertainty quality.}
Throughout the series of experiments we conducted list in Table~\ref{Tab:backbone_nyu}~and \ref{Tab:backbone_kitti}, from ResNet~\cite{he2016deep_resnet}, DenseNet~\cite{huang2017densely} to Swin transformer~\cite{liu2021swin}, we can easily tell that with better backbone models, we can expect improvements on both model accuracy and uncertainty, which is consistent with the observation in~\cite{RevisitingCalib_minderer_2021_nips}, suggesting that architecture is a major determinant of calibration properties.

\subsection{Discussion}
\label{sec:discussion}
\noindent\textit{Combinations with existing
methods:}
We show in Table~\ref{Tab:combine_kitti} that by utilizing the probability distribution, our method can
be combined with existing solutions with
better uncertainty results. We measure the mean probability of multiple predictions, 
using~\cite{deepensemble} (DE), \cite{gal2016dropout} (MCD) and \cite{mi2022training} (Noi). We see improvements in both depth prediction and uncertainty quality.
For GrUMO~\cite{Hornauer2022GradientbasedUF}, we obtain uncertainty from gradients of squared error between predictions of horizontally-flipped input pairs, while
better uncertainty is obtained by adding an extra term of difference between predicted probabilities of the horizontally-flipped input pairs. 

\begin{table}[t]
\centering
\footnotesize
\renewcommand{\arraystretch}{1.2}
\renewcommand{\tabcolsep}{0.1mm}
\caption{ Our methods combined with others, evaluated on KITTI. }
\resizebox{\linewidth}{!}{
 \begin{tabular}{r|c|cc|cc|cc|c}\toprule
    Method~ & ~B~ & ~RMSE$\downarrow$~ & ~AUSE$\downarrow$~ & ~~~Rel$\downarrow$~~~ & ~AUSE$\downarrow$~ & ~~~$\delta_1\uparrow$~~~ & \scriptsize{AUROC}$\uparrow$& ~~SCC$\uparrow$~~ \\\midrule
    Ours-C~ & R &3.4090 & 0.3448 & 0.0970 & 0.0195 & 0.8952 & 0.8609 & 0.6862 \\ \midrule

    DE~\cite{deepensemble} & \tr{R} & \tr{3.3819} & 0.7181 & \tr{0.0946} & 0.0279 & \tr{0.8994} & 0.8020 & 0.5419\\
                    +Ours~ &        &             & 0.3384 &             & 0.0195 &             & 0.8590 & 0.6817\\ \midrule

    MCD~\cite{gal2016dropout} & \tr{R} & \tr{3.3825} & 0.6652 & \tr{0.1004} & 0.0296 & \tr{0.8932} & 0.8068 & 0.5817\\
                       +Ours~ &        &             & 0.3500 &             & 0.0203 &             & 0.8620 & \b{0.6951}\\ \midrule

    Noi~\cite{mi2022training} & \tr{R} & \tr{\b{3.3298}} & 0.6044 & \tr{\b{0.0909}} & 0.0230 & \tr{\b{0.9057}} & 0.8416 & 0.5907\\
                       +Ours~ &        &             & \b{0.3325} &             & \b{0.0190} &             & \b{0.8670} & 0.6806\\ \midrule
    GrUMO~\cite{Hornauer2022GradientbasedUF} & \tr{R}&      \tr{3.4090} &    0.9491 &    \tr{0.0970} &    0.0316 &    \tr{0.8952} &    0.7789 & 0.5030\\ 
    + Ours~& & & 0.9473 & & 0.0315 & & 0.7743 & 0.5093 \\ \bottomrule
\end{tabular}}
\label{Tab:combine_kitti}
\end{table}

\begin{figure}[t]
\centering
\begin{tabular}{ c@{ } c@{ } c@{ } c@{ }}
    {\includegraphics[width=0.25\linewidth, trim=120 20 0 20, clip]{figs/volume_rend/00185_colors.png}}&
    {\includegraphics[width=0.25\linewidth, trim=120 20 0 20, clip]{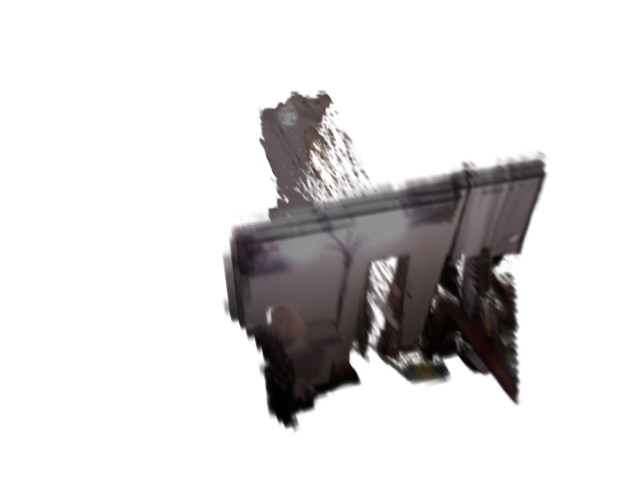}}&
    {\includegraphics[width=0.25\linewidth, trim=120 20 0 20, clip]{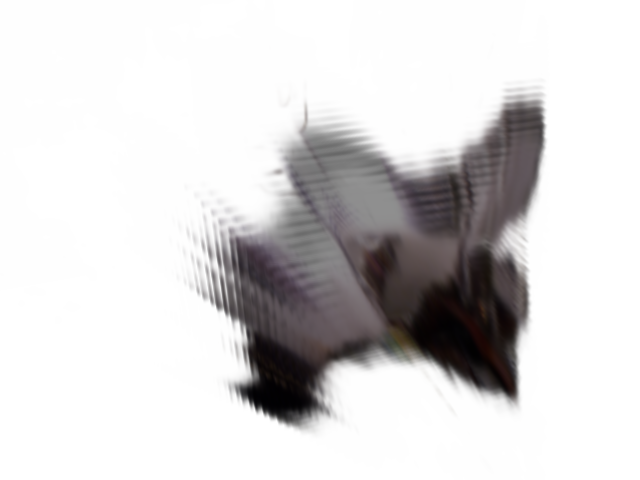}}&
    {\includegraphics[width=0.25\linewidth, trim=120 20 0 20, clip]{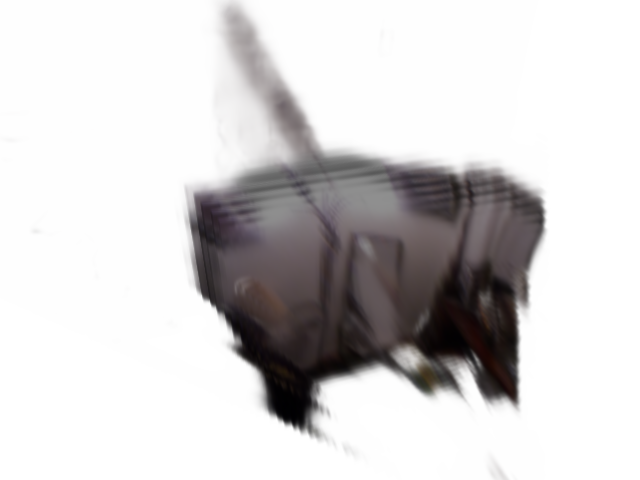}}\\
    
    {\includegraphics[width=0.24\linewidth, trim=110 10 0 30, clip]{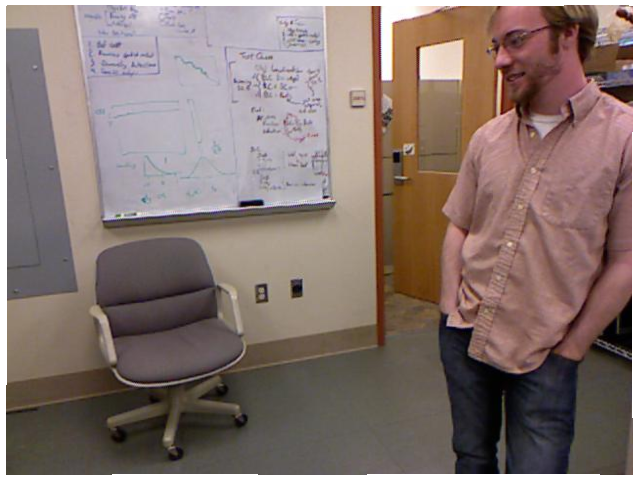}}&
    {\includegraphics[width=0.24\linewidth, trim=110 10 0 30, clip]{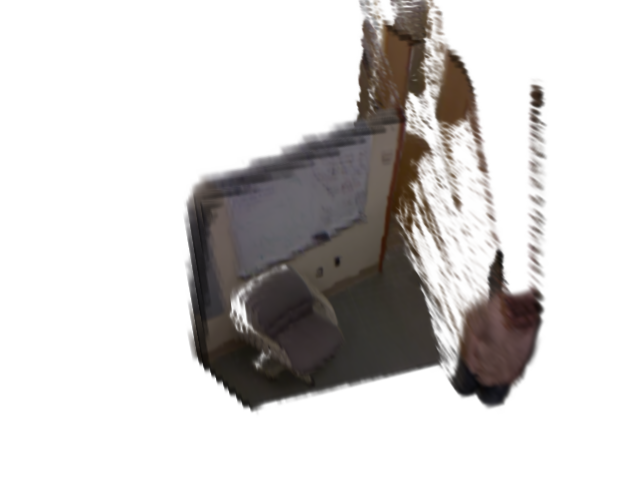}}&
    {\includegraphics[width=0.24\linewidth, trim=110 10 0 30, clip]{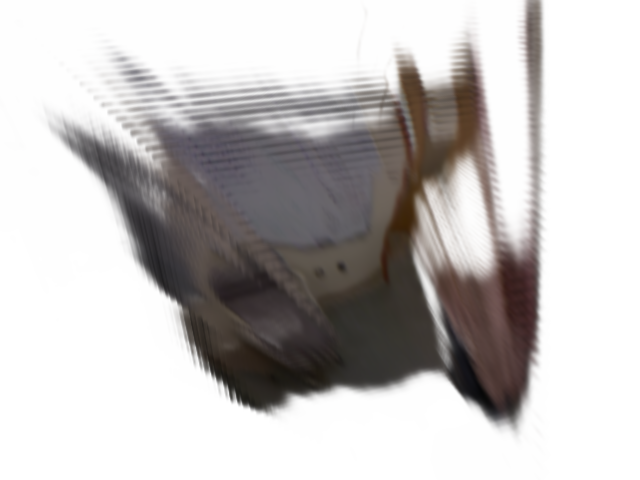}}&
    {\includegraphics[width=0.24\linewidth, trim=110 10 0 30, clip]{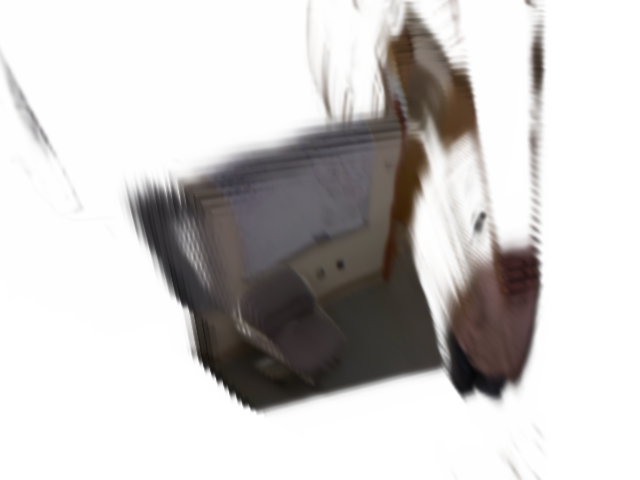}}\\
        
    {\includegraphics[width=0.24\linewidth, trim=100 0 0 110, clip]{figs/volume_rend/01163_colors.png}}&
    {\includegraphics[width=0.24\linewidth, trim=100 0 0 110, clip]{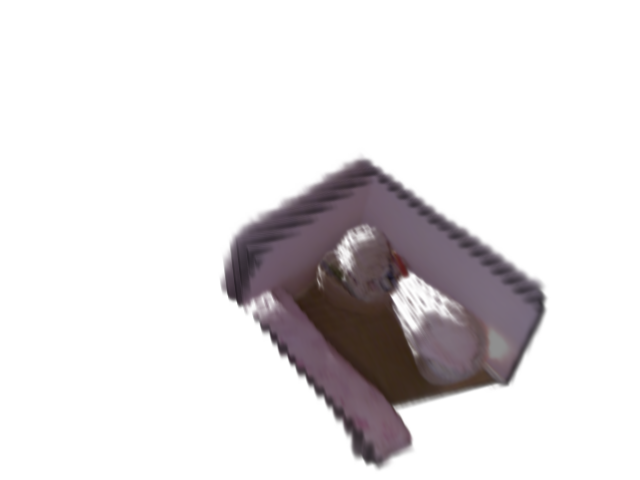}}&
    {\includegraphics[width=0.24\linewidth, trim=100 0 0 110, clip]{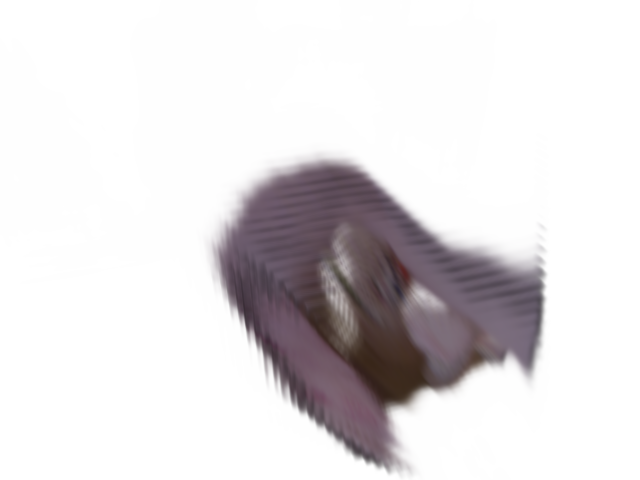}}&
    {\includegraphics[width=0.24\linewidth, trim=100 0 0 110, clip]{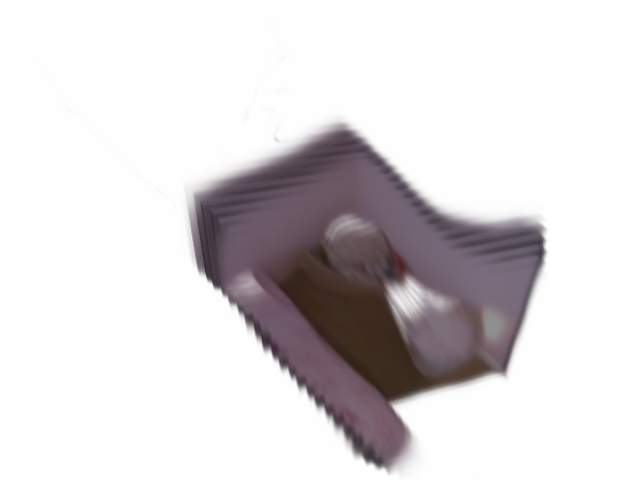}}\\
    
    {\includegraphics[width=0.24\linewidth]{figs/volume_rend/01249_colors.png}}&
    {\includegraphics[width=0.24\linewidth]{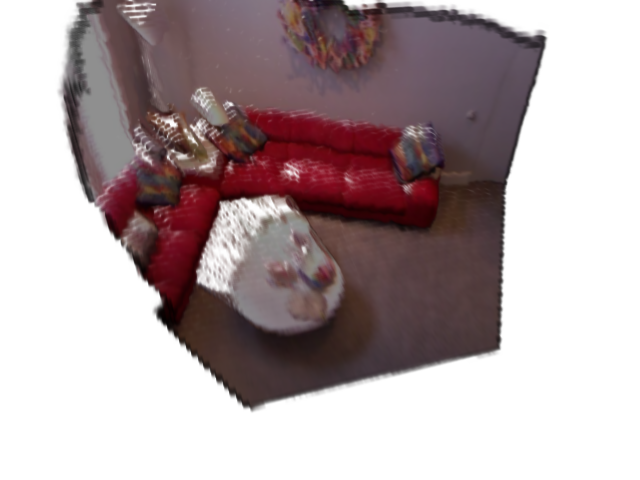}}&
    {\includegraphics[width=0.24\linewidth]{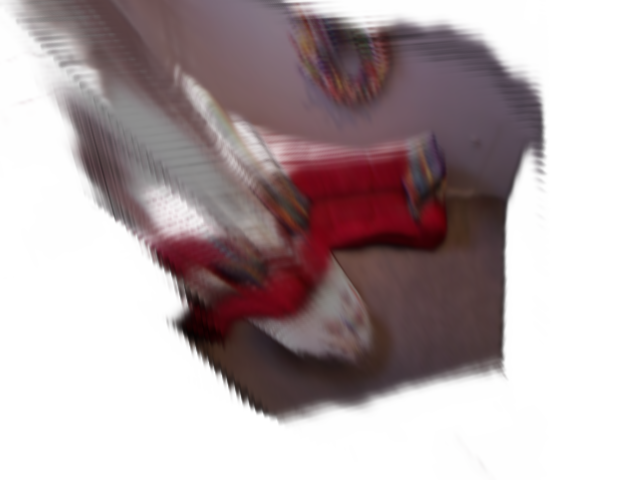}}&
    {\includegraphics[width=0.24\linewidth]{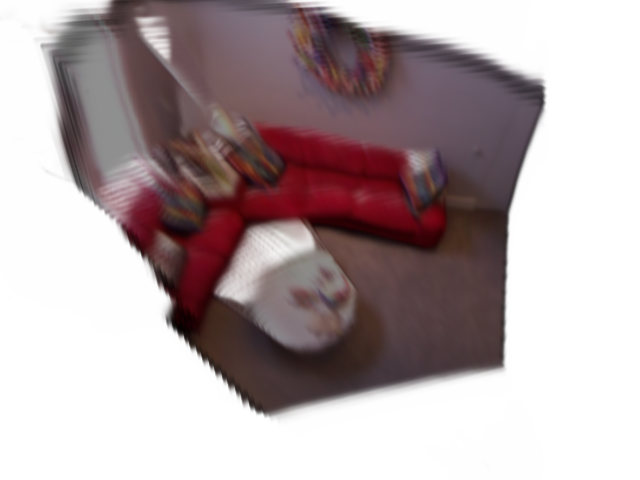}}\\
    
    {\includegraphics[width=0.24\linewidth]{figs/volume_rend/01253_colors.png}}&
    {\includegraphics[width=0.24\linewidth]{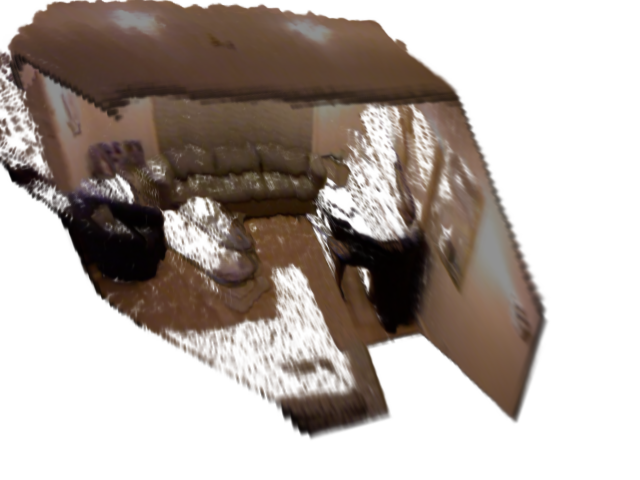}}&
    {\includegraphics[width=0.24\linewidth]{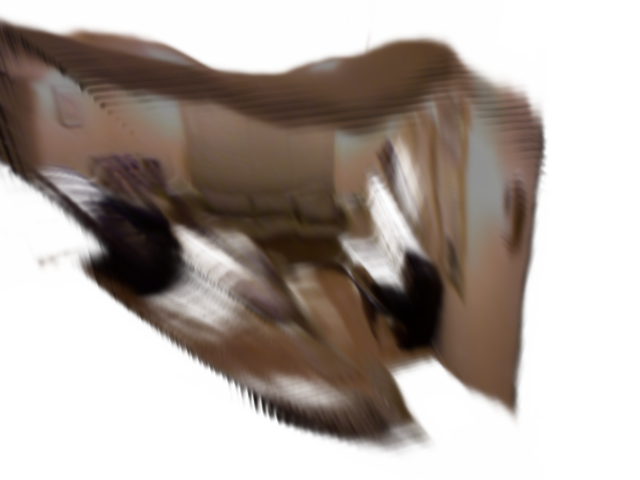}}&
    {\includegraphics[width=0.24\linewidth]{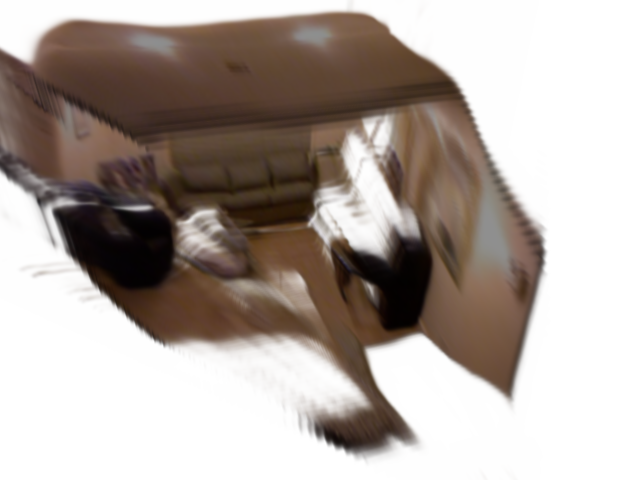}}\\
    
    {\includegraphics[width=0.24\linewidth, trim=140 0 0 40, clip]{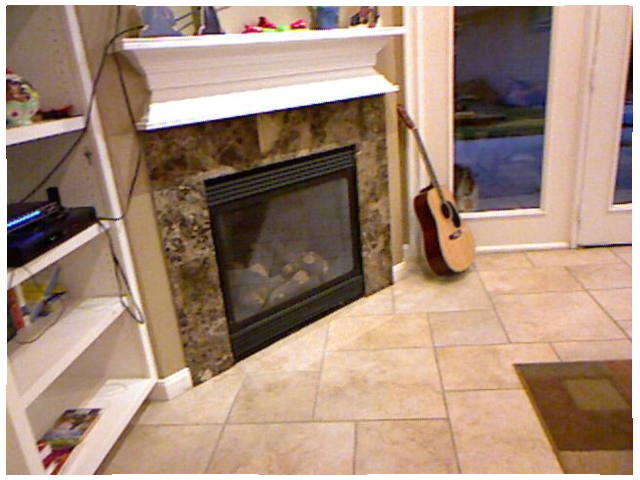}}&
    {\includegraphics[width=0.24\linewidth, trim=140 0 0 40, clip]{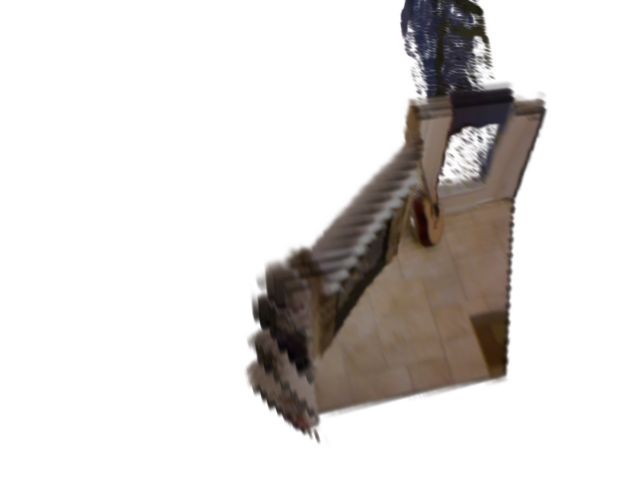}}&
    {\includegraphics[width=0.24\linewidth, trim=140 0 0 40, clip]{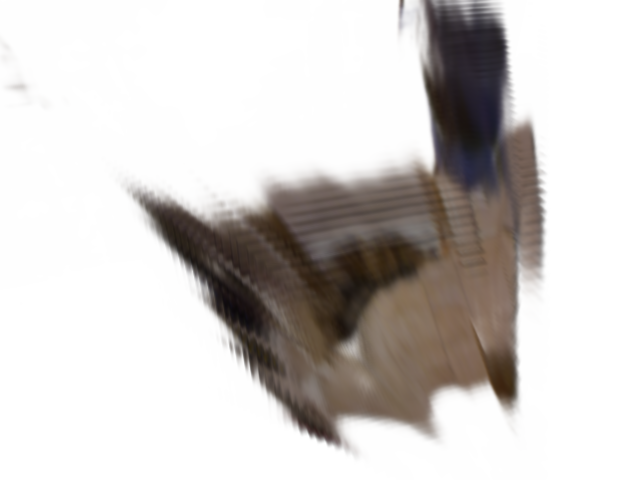}}&
    {\includegraphics[width=0.24\linewidth, trim=140 0 0 40, clip]{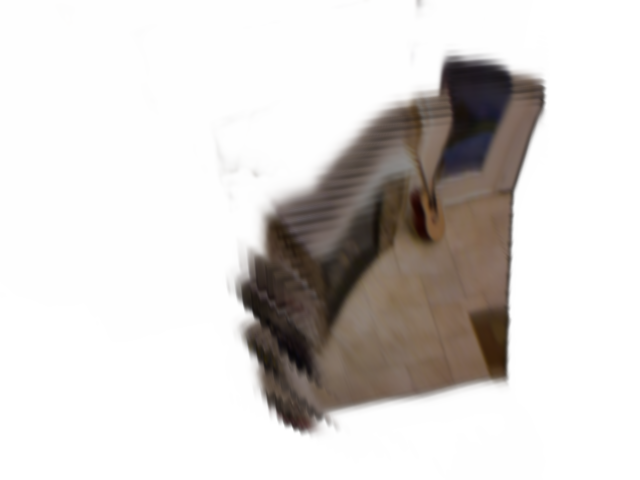}}\\
    \small Image & \small GT & \small Ours-C,R & \small Ours-C,S \\
  \end{tabular}
\caption{Volume rendering results of GT and the depth probability volume of our methods.}
\label{fig:uncert_vis}
\end{figure}

\noindent\textit{Significance of our solution:} Based on the proposed deterministic uncertainty generation method 
Eq.~\ref{ue_defination}
, our loss regularization (Eq.~\ref{loss_reg}) is 
a simple and straightforward way
to model the ordinal-sensitive nature of MDE. We discuss the limitations of existing uncertainty measures, \ie,~AUSE, and introduce an alternative uncertainty measure with Spearman correlation coefficients in Sec.~\ref{sec:sparsification flaw}. Our 3D depth probability volume for uncertainty visualization 
(Fig.~\ref{fig:uncert_vis})
is expressive in explainable uncertainty.

\section{Conclusion}
\label{sec:conclusion}
Given the \enquote{ordinal-sensitive} nature of MDE, we introduced a simple and effective method with reliable deterministic uncertainty generated by utilizing the probability distributions. We have demonstrated that by adding extra regularization terms, we can improve model uncertainty without adding computational overhead. Our method is intuitive and with high extensibility, that can be combined with other uncertainty measuring methods to further improve the reliability of uncertainty. We also discuss the limitation of the wildly-used Sparsification error as an uncertainty measure, and introduce Spearman correlation coefficients as an alternative and more suitable uncertainty measure to evaluate the uncertainty quality of different models. Our 3D visualization of the depth probability offers more expressive illustrations of the generated uncertainty.


{\small
\bibliographystyle{ieee_fullname}
\bibliography{Depth_Uncertainty}
}
\clearpage
\makeatletter
\AfterEndEnvironment{algorithm}{\let\@algcomment\relax}
\AtEndEnvironment{algorithm}{\kern2pt\hrule\relax\vskip3pt\@algcomment}
\let\@algcomment\relax
\renewcommand\fs@ruled{\def\@fs@cfont{\bfseries}\let\@fs@capt\floatc@ruled
  \def\@fs@pre{\hrule height.8pt depth0pt \kern2pt}%
  \def\@fs@post{}%
  \def\@fs@mid{\kern2pt\hrule\kern2pt}%
  \let\@fs@iftopcapt\iftrue}
\makeatother



\def\Jing#1{{\color{magenta}{\bf [Jing:} {\it{#1}}{\bf ]}}}
\def\YC#1{{\color{blue}{\bf [Yuchao:} {\it{#1}}{\bf ]}}}
\def\Mochu#1{{\color{orange}{\bf [Mochu:} {\it{#1}}{\bf ]}}}
\def\NB#1{{\color{green}{\bf [NB:} {\it{#1}}{\bf ]}}}

\def\tr#1{\multirow{2}{*}{#1}} 
\def\tc#1{\multicolumn{2}{c|}{#1}} 
\def\fc#1{\multicolumn{4}{c|}{#1}} 
\def\da{$\downarrow$}
\def\ua{$\uparrow$}

\def\rf{\cellcolor[HTML]{B4C7E7}} 
\def\rs{\cellcolor[HTML]{D2DDF1}} 
\def\rt{\cellcolor[HTML]{E9EEF8}} 
\def\tr#1{\multirow{2}{*}{#1}} 
\def\b#1{\textbf{#1}} 
\def\wl#1{\multicolumn{1}{c}{\begin{tabular}[c]{@{}c@{}}{#1}\end{tabular}}} 


\def\iccvPaperID{5676} 
\def\httilde{\mbox{\tt\raisebox{-.5ex}{\symbol{126}}}}

\ificcvfinal\pagestyle{empty}\fi




\section{Model Details}

\begin{figure}[hbtp]
\centering
\includegraphics[width=0.78\linewidth]{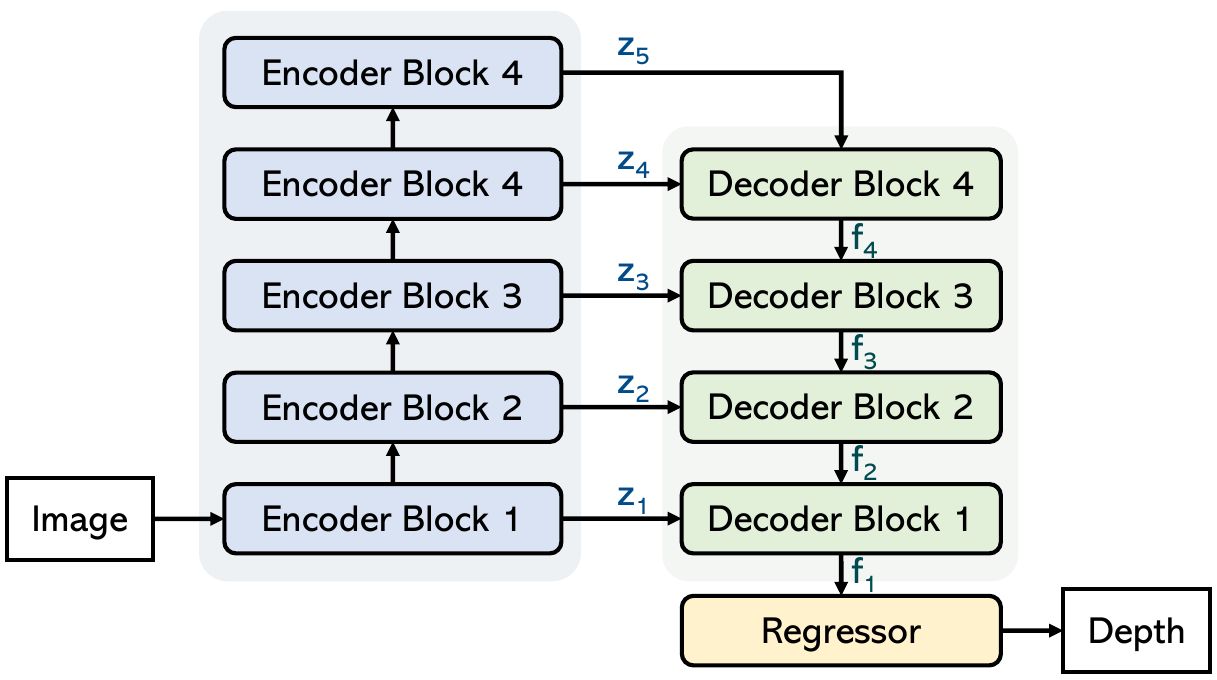}
\caption{The architecture of our models.}
\label{fig:en_decoder}
\end{figure}

\begin{figure}[bthp!]
\centering
\includegraphics[width=0.85\linewidth]{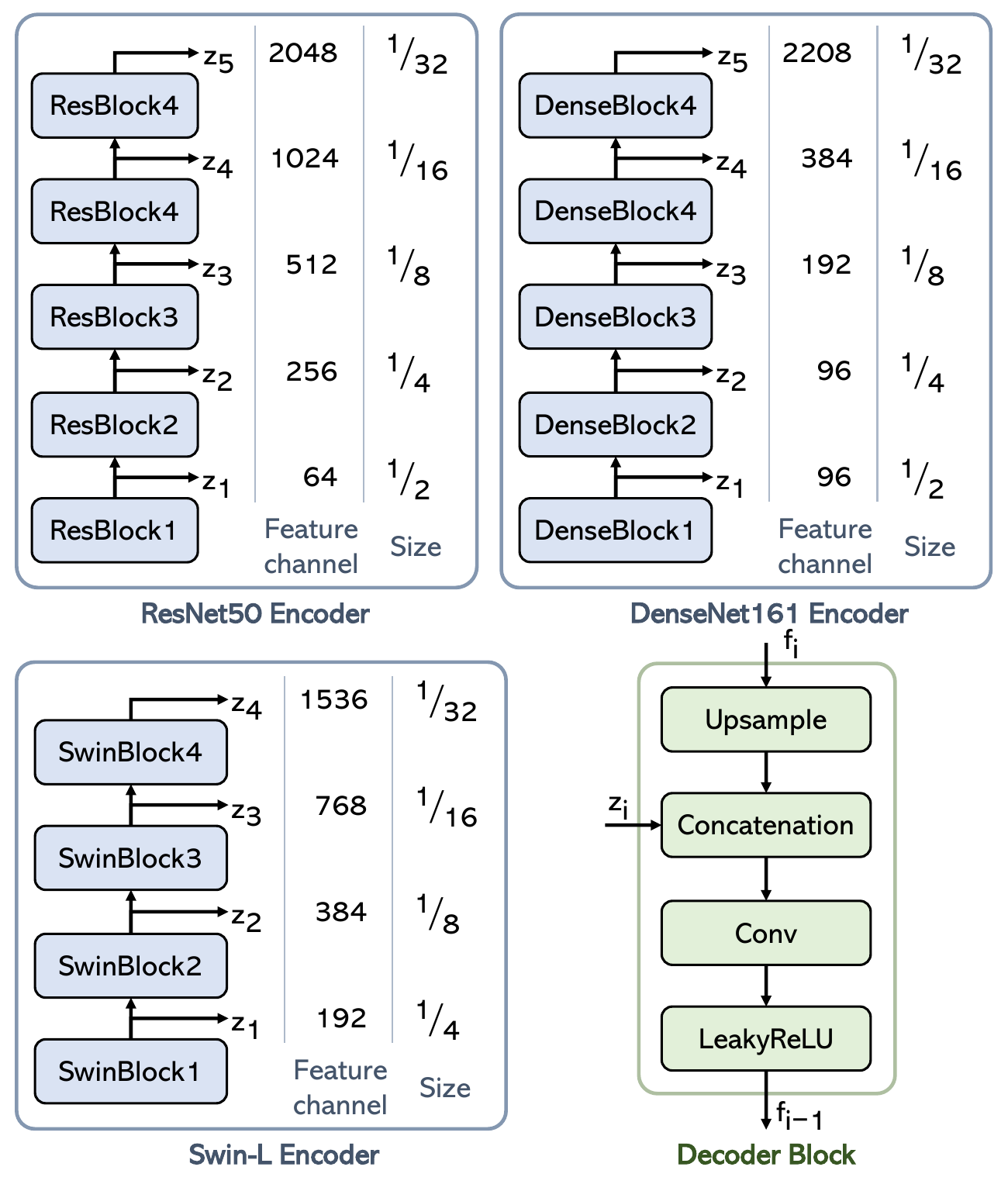}
\caption{The details of encoder and decoder components.}
\label{fig:model_architecture}
\end{figure}

We show the basic structure of our MDE models in Figure~\ref{fig:en_decoder}, and detailed architectures of encoder and decoder components in Figure~\ref{fig:model_architecture}.

\begin{table*}[t!]
\centering
\footnotesize
\renewcommand{\arraystretch}{1.1}
\renewcommand{\tabcolsep}{2.5mm}
\caption{ Accuracy metrics on NYU. }
 \begin{tabular}{r|c |ccccc|ccc}\toprule
     Method~ & ~Backbone~ & ~~~~Rel\da~~~~ & ~~~log10\da~~~ & ~RMSE\da~ & ~SqRel\da~ & ~logRMS\da~ & ~~~~$\delta_1$\ua~~~~& ~~~~$\delta_2 \uparrow$~~~~ & ~~~~$\delta_3 \uparrow$~~~~  \\\midrule
     DE~\cite{deepensemble}~ & ResNet50 &  0.2051 & 0.0863 & 0.6936 & 0.1948 & 0.2547 & 0.6735 & 0.9006 & 0.9685 \\

     MCD~\cite{gal2016dropout}~ & ResNet50 &  0.2126 & 0.0881 & 0.7054 & 0.2068 & 0.2580 & 0.6660 & 0.8961 & 0.9655 \\

     Noise~\cite{mi2022training}~ & ResNet50 & 0.1978 & 0.0847 & 0.6814 & 0.1836 & 0.2509 & 0.6827 & 0.9037 & 0.9695 \\

     LDU~\cite{latent_Discriminant_deterministic_uncertainty}~ & ~DenseNet161~ &  0.1132 & 0.0483 & 0.4015 & 0.0662 & 0.1447 & 0.8739 & 0.9789 & 0.9949 \\

     SLU~\cite{yu2021slurp}~ & ~DenseNet161~ &  0.1100 & 0.0471 & 0.3970 & 0.0672 & 0.1425 & 0.8845 & 0.9774 & 0.9943 \\
 \midrule




Ours-C~ & ResNet50 & 0.2132 & 0.0903 & 0.7280 & 0.2103 & 0.2666 & 0.6553 & 0.8888 & 0.9639 \\
Ours-C~ & ~DenseNet161~ & 0.1289 & 0.0556 & 0.4811 & 0.0905 & 0.1698 & 0.8361 & 0.9652 & 0.9910 \\
Ours-C~ & Swin-L & 0.1124 & 0.0480 & 0.4275 & 0.0756 & 0.1484 & 0.8813 & 0.9752 & 0.9914 \\ \midrule
Ours-R~ & ResNet50 & 0.2180 & 0.0915 & 0.7283 & 0.2160 & 0.2682 & 0.6465 & 0.8867 & 0.9635 \\
Ours-R~ & ~DenseNet161~ & 0.1310 & 0.0562 & 0.4787 & 0.0899 & 0.1699 & 0.8347 & 0.9653 & 0.9911 \\
Ours-R~ & Swin-L & 0.1118 & 0.0479 & 0.4196 & 0.0704 & 0.1455 & 0.8794 & 0.9765 & 0.9942 \\  \bottomrule
\end{tabular}
\label{Tab:nyu_acc_full}
\end{table*}

\begin{table*}[t!]
\centering
\footnotesize
\renewcommand{\arraystretch}{1.1}
\renewcommand{\tabcolsep}{2.5mm}
\caption{ Accuracy metrics on KITTI. }
 \begin{tabular}{r|c |ccccc|ccc}\toprule
     Method~ &  ~Backbone~  & ~~~~Rel\da~~~~ & ~~~log10\da~~~ & ~RMSE\da~ & ~SqRel\da~ & ~logRMS\da~ & ~~~~$\delta_1$\ua~~~~& ~~~~$\delta_2 \uparrow$~~~~ & ~~~~$\delta_3 \uparrow$~~~~  \\\midrule
    DE~\cite{deepensemble}~ & ResNet50 &  0.0946 & 0.0404 & 3.3819 & 0.4852 & 0.1447 & 0.8994 & 0.9765 & 0.9927 \\ 
    MCD~\cite{gal2016dropout}~ & ResNet50 &   0.1005 & 0.0418 & 3.3845 & 0.5136 & 0.1496 & 0.8932 & 0.9723 & 0.9910 \\ 
    Noise~\cite{mi2022training}~ & ResNet50 & 0.0909 & 0.0393 & 3.3296 & 0.4418 & 0.1397 & 0.9058 & 0.9780 & 0.9937 \\ 
    LDU~\cite{latent_Discriminant_deterministic_uncertainty}~ & ~DenseNet161~ &  0.0674 & 0.0298 & 3.0763 & 0.3545 & 0.1080 & 0.9457 & 0.9899 & 0.9975 \\ 
    SLU~\cite{yu2021slurp}~ & ~DenseNet161~ & 0.0639 & 0.0282 & 2.9466 & 0.2783 & 0.1032 & 0.9486 & 0.9911 & 0.9977 \\ \midrule

Ours-C~ & ResNet50 & 0.0970 & 0.0415 & 3.4090 & 0.5066 & 0.1481 & 0.8952 & 0.9738 & 0.9918 \\
Ours-C~ & ~DenseNet161~ & 0.0632 & 0.0275 & 2.5502 & 0.2331 & 0.0995 & 0.9538 & 0.9924 & 0.9981 \\
Ours-C~ & Swin-L & 0.0571 & 0.0251 & 2.4095 & 0.2008 & 0.0895 & 0.9651 & 0.9950 & 0.9987 \\ \midrule
Ours-R~ & ResNet50 & 0.0966 & 0.0415 & 3.4172 & 0.4854 & 0.1482 & 0.8940 & 0.9731 & 0.9918 \\
Ours-R~ & ~DenseNet161~ & 0.0641 & 0.0281 & 2.5716 & 0.2378 & 0.1006 & 0.9529 & 0.9919 & 0.9979 \\
Ours-R~ & Swin-L & 0.0571 & 0.0250 & 2.3758 & 0.1956 & 0.0889 & 0.9655 & 0.9951 & 0.9988 \\

         \bottomrule
\end{tabular}
\label{Tab:kitti_acc_full}

\end{table*}

\begin{table*}[t!]
\centering
\footnotesize
\renewcommand{\arraystretch}{1.1}
\renewcommand{\tabcolsep}{1.3mm}
\caption{ Uncertainty metrics on NYU. }
 \begin{tabular}{r|c|cc|cc|cccc|c}\toprule
     \tr{Method}~ & ~\tr{Backbone}~ & \tc{RMSE} & \tc{Rel} & \fc{$\delta_1$} & ~~~\tr{SCC\ua}~~~ \\
                  &          & ~~AUSE\da~~ & ~~AURG\ua~~ & ~~AUSE\da~~ & ~~AURG\ua~~ & ~~AUSE\da~~ & ~~AURG\ua~~ & ~~FPR95\ua~~ & ~~AUROC\ua~~ & \\ \midrule
     DE~\cite{deepensemble}~ & ResNet50 & 0.3165 & 0.1233 & 0.0954 &  0.0231 & 0.1956 &  0.0426 & 0.8993 & 0.5933 & 0.1914 \\

     MCD~\cite{gal2016dropout}~ & ResNet50 & 0.3223 & 0.1229 & 0.1009 &  0.0209 & 0.2013 &  0.0391 & 0.8978 & 0.5823 & 0.2013 \\

     Noise~\cite{mi2022training}~ & ResNet50 & 0.2877 & 0.1468 & 0.0875 &  0.0256 & 0.1775 &  0.0541 & 0.8507 & 0.6186 & 0.2399 \\

     LDU~\cite{latent_Discriminant_deterministic_uncertainty}~ & ~DenseNet161~ & 0.1257 & 0.1327 & 0.0635 & -0.0009 & 0.1093 & -0.0059 & 0.7541 & 0.5621 & 0.3117 \\

     SLU~\cite{yu2021slurp}~ & ~DenseNet161~ &  0.1144 & 0.1432 & 0.0501 &  0.0116 & 0.0739 &  0.0214 & 0.6447 & 0.6820 & 0.3406 \\
 \midrule




Ours-C~ & ResNet50 & 0.2781 & 0.1867 & 0.0894 & 0.0338 & 0.1813 & 0.0666 & 0.7738 & 0.6349 & 0.2525 \\
Ours-C~ & ~DenseNet161~ & 0.1534 & 0.1663 & 0.0504 & 0.0238 & 0.0841 & 0.0509 & 0.6050 & 0.7217 & 0.3099 \\
Ours-C~ & Swin-L & 0.1257 & 0.1591 & 0.0420 & 0.0217 & 0.0539 & 0.0455 & 0.5411 & 0.7656 & 0.3284 \\ \midrule
Ours-R~ & ResNet50 & 0.2767 & 0.1822 & 0.0911 & 0.0342 & 0.1894 & 0.0635 & 0.7804 & 0.6257 & 0.2480 \\
Ours-R~ & ~DenseNet161~ & 0.1488 & 0.1642 & 0.0498 & 0.0245 & 0.0796 & 0.0547 & 0.5803 & 0.7324 & 0.3124 \\
Ours-R~ & Swin-L & 0.1189 & 0.1568 & 0.0425 & 0.0200 & 0.0561 & 0.0441 & 0.5167 & 0.7639 & 0.3281 \\

         \bottomrule
\end{tabular}
\label{Tab:nyu_unc_full}
\end{table*}

\begin{table*}[t!]
\centering
\footnotesize
\renewcommand{\arraystretch}{1.05}
\renewcommand{\tabcolsep}{1.5mm}
\caption{ Uncertainty metrics on KITTI. }
 \begin{tabular}{r|c|cc|cc|cccc|c}\toprule
     \tr{Method}~ & ~\tr{Backbone}~ & \tc{RMSE} & \tc{Rel} & \fc{$\delta_1$} & ~~\tr{SCC\ua}~~ \\
                  &          & ~~AUSE\da~~ & ~~AURG\ua~~ & ~~AUSE\da~~ & ~~AURG\ua~~ & ~~AUSE\da~~ & ~~AURG\ua~~ & ~~FPR95\ua~~ & ~~AUROC\ua~~ & \\ \midrule
    DE~\cite{deepensemble}~ & ResNet50 &  0.7181 & 2.1261 & 0.0279 & 0.0357 & 0.0311 & 0.0604 & 0.5551 & 0.8020 & 0.5419 \\
    MCD~\cite{gal2016dropout}~ & ResNet50 &   0.6574 & 2.1774 & 0.0295 & 0.0386 & 0.0327 & 0.0629 & 0.5470 & 0.8083 & 0.5818 \\
    Noise~\cite{mi2022training}~ & ResNet50 & 0.6025 & 2.1991 & 0.0227 & 0.0373 & 0.0216 & 0.0640 & 0.4355 & 0.8453 & 0.5948 \\
    LDU~\cite{latent_Discriminant_deterministic_uncertainty}~ & ~DenseNet161~ &  0.3256 & 2.3165 & 0.0200 & 0.0235 & 0.0155 & 0.0340 & 0.4108 & 0.8433 & 0.6704 \\
    SLU~\cite{yu2021slurp}~ & ~DenseNet161~ &  0.2662 & 2.2709 & 0.0140 & 0.0279 & 0.0089 & 0.0390 & 0.2652 & 0.8948 & 0.7008 \\ \midrule

Ours-C~ & ResNet50 & 0.3448 & 2.5138 & 0.0195 & 0.0454 & 0.0201 & 0.0740 & 0.3486 & 0.8609 & 0.6862 \\
Ours-C~ & ~DenseNet161~ & 0.2352 & 1.9396 & 0.0139 & 0.0274 & 0.0080 & 0.0356 & 0.2501 & 0.8985 & 0.6779 \\
Ours-C~ & Swin-L & 0.2199 & 1.8361 & 0.0128 & 0.0235 & 0.0052 & 0.0282 & 0.2325 & 0.9109 & 0.6735 \\ \midrule
Ours-R~ & ResNet50 & 0.3408 & 2.5288 & 0.0193 & 0.0456 & 0.0195 & 0.0764 & 0.3445 & 0.8657 & 0.6909 \\
Ours-R~ & ~DenseNet161~ & 0.2369 & 1.9494 & 0.0138 & 0.0276 & 0.0082 & 0.0358 & 0.2530 & 0.8988 & 0.6778 \\
Ours-R~ & Swin-L & 0.2072 & 1.8148 & 0.0127 & 0.0236 & 0.0048 & 0.0283 & 0.2253 & 0.9125 & 0.6790 \\

         \bottomrule
\end{tabular}
\label{Tab:kitti_unc_full}
\end{table*}
\section{Full Metrics}

Due to the limited space of the paper, we were not able to present the full metrics. Here we show full metrics for all models. In Table~\ref{Tab:nyu_acc_full} and \ref{Tab:kitti_acc_full} we show full accuracy metrics on NYU and KITTI. In Table~\ref{Tab:nyu_unc_full} and \ref{Tab:kitti_unc_full}, we show full accuracy metrics on NYU and KITTI. Readers can refer to~\cite{eigen2014depth} and~\cite{ilg2018uncertainty} for details about accuracy and uncertainty metrics.


Readers may notice the difference in accuracy among models adopting DenseNet161 as backbone, especially in the NYU~\cite{silberman2012indoor} dataset. This is because that our models adopt a simpler decoder from~\cite{godard2019digging}, while LDU~\cite{latent_Discriminant_deterministic_uncertainty} and SLU~\cite{yu2021slurp} adopt a more complex decoder from BTS~\cite{lee2019big}, which utilizes planar
guidance.

\section{Uncertainty Visualization}

\subsection{Basic Principles}

Conventional uncertainty visualization illustrates uncertainty estimation as a 2D map. Since a classification MDE model estimates probability that the depth of a pixel falls into a certain depth interval, which can be interpreted as the possibility that there exists a physical point at a certain location. Intuitively, such an interpretation can be further extended to regarding the possibility as the opacity of points in the viewing frustum, and that the color of points are shared along the same viewing ray;
that is, we assign the same color to all 3D points that project to the same location on the 2D image using the color of that pixel.
Given this setting, it is natural to visualize the probability volume using volume rendering methods. 

The benefits of such a visualization are two-fold. Firstly, compared to the 2D visualization of uncertainty, the volume rendered probability volume not only tells how uncertain the prediction is, it also tells where the object might be. Secondly, compared to the 3D visualization of depth, the 3D visualization of probabilities presents richer information of model predictions, \ie a distribution rather than a single value, so that it can handle predictions of multiple possible depth values.

\begin{figure}[t!]
\centering
\includegraphics[width=0.9\linewidth, trim=0 10 0 23, clip]{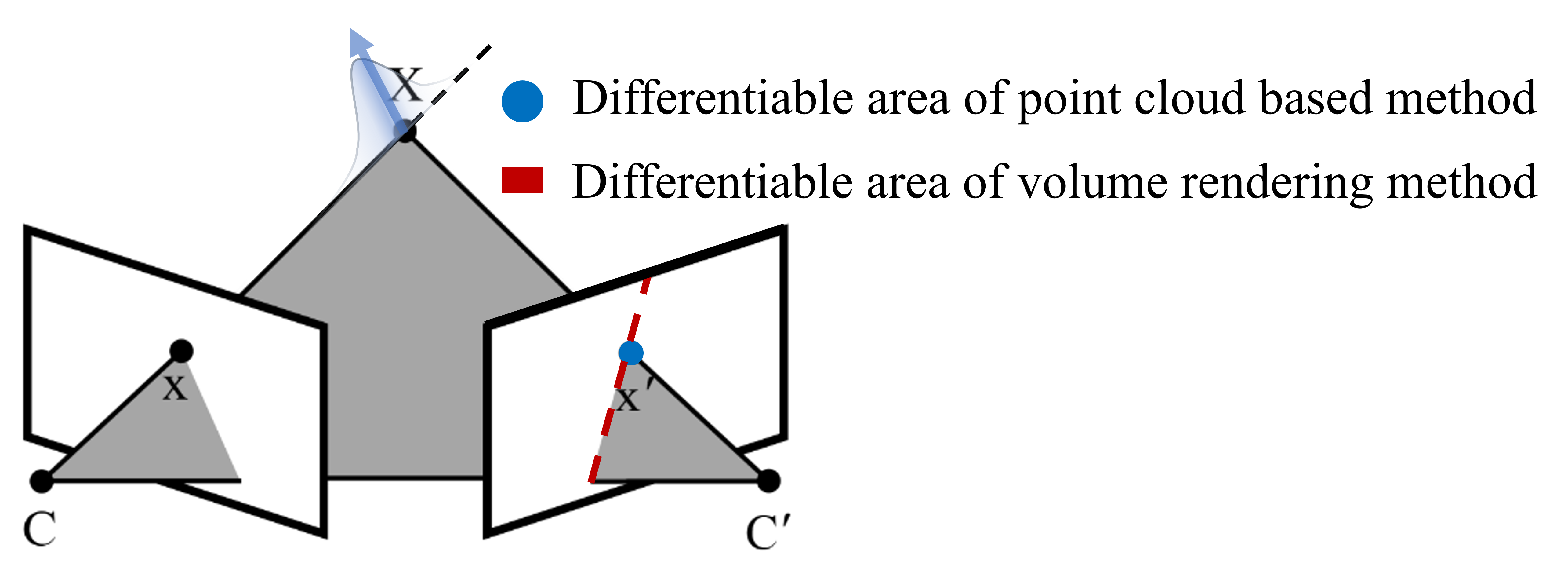}
\vspace{-1mm}
\caption{A comparison of differentiability between point cloud based method and volume rendering method.}
\label{fig:differentiable}
\end{figure}

Further, this representation has better differentiability than point cloud based methods.
In Figure~\ref{fig:differentiable}, for a point $\mathbf x$ in the first view and its estimated depth value, point cloud based method generate a 3D point $\mathbf X$ and project it to the second view, resulting a point $\mathbf x'$; this makes the depth estimation only differentiable within the area of $\mathbf x'$. In contrast, the entire epipolar line is differentiable with respect to the depth probability estimations in the second view if using volume rendering methods.


\subsection{Technical Details of Volume Rendering}

\begin{figure}[t!]
\centering
\includegraphics[width=0.9\linewidth, trim=0 15 0 10, clip]{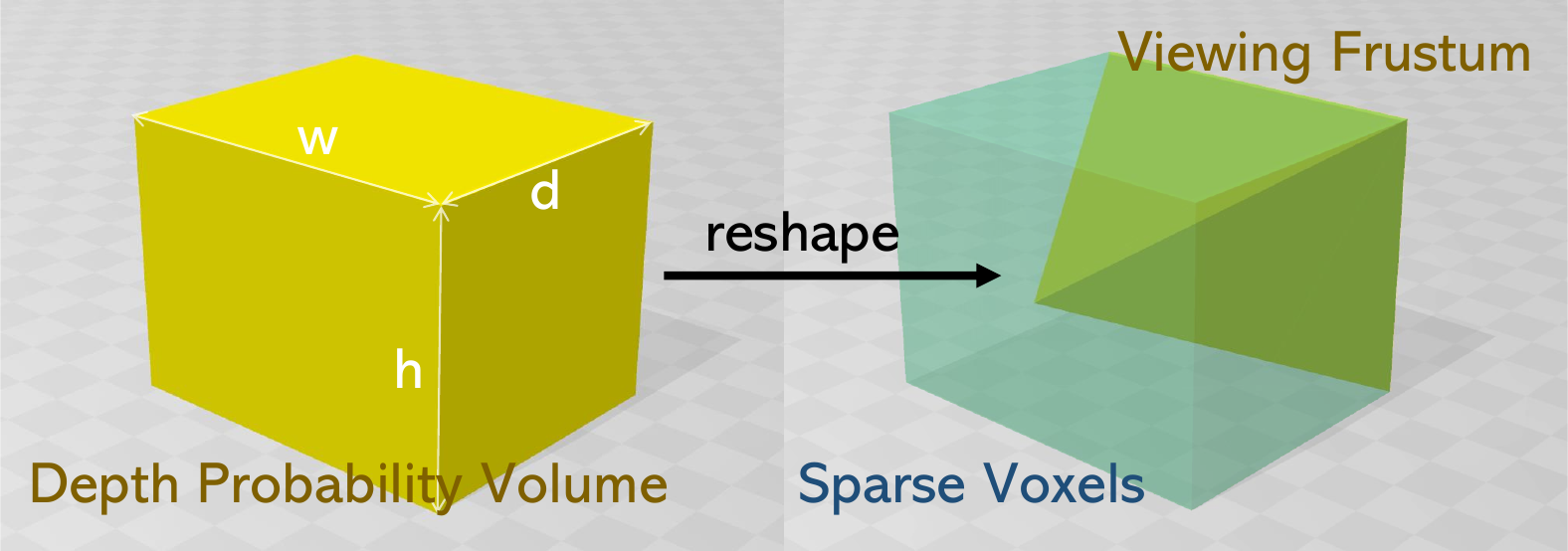}
\caption{Forming the depth probability volume into sparse voxels.}
\label{fig:forming_voxels}
\end{figure}

In Figure~\ref{fig:forming_voxels} we show how to form the estimated depth probability volume into sparse voxels.

The locations of values in the depth probability volume do not correspond to the points uniformly located in 3D space. In fact, the depth probability volume can only predict the depth value of the objects falling into the viewing frustum. In order to visualize the estimated probability, we need to reshape the depth probability volume in the shape of cube into the frustum. The depth hypothesis plane with the smallest depth value (1e-3 in our setting) collapses into a infinitesimal point, and the shape of the viewing frustum is determined considering an extra parameter, the focal length. 

We bilinearly assign weights of the ground truth value to the nearest depth probability volume, while the reshaping of the depth probability volume to the sparse voxels adopts trilinear interpolation.

With the above reshaping process, the probability values are regarded as opacity values stored in the sparse voxels, and they geometrically correspond to the 3D shape of the estimated scene. We then use volume rendering for sparse voxels to provide the uncertainty visualization.
The volume rendering framework for sparse voxels is from~\cite{yu2021plenoxels}.




\end{document}